%% file: main.tex
\documentclass[nohyperref]{article}

\usepackage{microtype}
\usepackage{graphicx}
\usepackage{subcaption}
\usepackage{booktabs} 
\usepackage[table]{xcolor}
\usepackage{makecell}

\usepackage{hyperref}

\usepackage[accepted]{icml2022}

\usepackage{comment}
\usepackage{amsmath}
\usepackage{makecell}
\usepackage{amssymb}
\usepackage{mathtools}
\usepackage{amsthm}
\usepackage{bbm}
\usepackage{enumitem}
\usepackage[capitalize,noabbrev]{cleveref}
\usepackage{kotex} 
\usepackage{multicol}
\usepackage{multirow}
\usepackage{csquotes}
\usepackage{anyfontsize}
\usepackage{wrapfig}

\usepackage{pifont}
\theoremstyle{plain}

\theoremstyle{definition}

\theoremstyle{remark}

\usepackage[textsize=tiny]{todonotes}

\icmltitlerunning{Risk Perspective Exploration in Distributional Reinforcement Learning}

\begin{document}

\twocolumn[
\icmltitle{Risk Perspective Exploration in Distributional Reinforcement Learning}




\begin{icmlauthorlist}
\icmlauthor{Jihwan Oh}{ai}
\icmlauthor{Joonkee Kim}{ai}
\icmlauthor{Se-Young Yun}{ai}
\end{icmlauthorlist}

\icmlaffiliation{ai}{Kim Jaechul Graduate School of Artificial Intelligence, KAIST, Seoul, South Korea}

\icmlcorrespondingauthor{Se-Young Yun}{yunseyoung@kaist.ac.kr}



\icmlkeywords{exploration, risk sensitiveness, reinforcement learning, distributional reinforcement learning}

\vskip 0.3in
]



\printAffiliationsAndNotice{}  

\begin{abstract}
Distributional reinforcement learning demonstrates state-of-the-art performance in continuous and discrete control settings with the features of variance and risk, which can be used to explore. However, the exploration method employing the risk property is hard to find, although numerous exploration methods in Distributional RL employ the variance of return distribution per action. In this paper, we present risk scheduling approaches that explore risk levels and optimistic behaviors from a risk perspective. We demonstrate the performance enhancement of the DMIX algorithm using risk scheduling in a multi-agent setting with comprehensive experiments.
\end{abstract}


\input{Manuscript/1_Introduction}
\input{Manuscript/2_Related_Works}
\input{Manuscript/3_Risk_scheduling}
\input{Manuscript/4_Experiments}
\input{Manuscript/5_Conclusion}
\input{Manuscript/6_Acknowledgements}




\nocite{langley00}

\bibliography{icml2022}
\bibliographystyle{icml2022}

\newpage
\appendix
\onecolumn

\input{Appendix/1_Distributional_Reinforcement_Learning}

\end{document}

%% file: Manuscript/1_Introduction.tex
\section{Introduction}
\label{sec:introduction}

Reinforcement learning \cite{sutton2018reinforcement} is a machine learning method that imitates the way humans learn to train models used in various domains such as robotics, autonomous driving, video games, economy, and industrial resource optimization. With one step further, the distributional perspective of deep reinforcement learning \cite{bellemare2017distributional, dabney2018distributional, dabney2018implicit}, has been highlighted with outperforming performance in Mujoco and Atari environment \citep{bellemare13arcade} compared to general reinforcement learning algorithms. In particular, one of the reasons why distributional RL shows state-of-the-art performance is that it utilizes exploration strategies with variances of return distributions which is a challenging issue in the reinforcement learning domain. \citet{mavrin2019distributional} and \citet{zhou2021non} propose exploration methods using the variance of a return distribution of distributional RL motivated by the existing reinforcement learning exploration strategy from \citet{burda2018exploration}, \citet{auer2002using}. However, exploration utilizing risk level, which determines cautious or daring behavior for agents, is hard to find despite having suitable property for exploration compared to methods using the variance in distributional RL. In this paper, we propose risk perspective exploration utilizing risk levels by simply scheduling it, enabling exploration of the risk sampling domain and less explored action space. We evaluate our method with Multi-Agent Reinforcement Learning (MARL) algorithm, which has a problem of challenging exploration. We employ the distributional MARL method, DMIX, a variation of DFAC \cite{sun2021dfac} that demonstrates state-of-the-art performance with variance and risk features but only implicitly employs a variance of return distribution as an exploration factor. By using risk as a factor of exploration in DMIX, we train agents based on the risk and show how the performance improves. We summarize our contributions as follows:
\begin{itemize}
    \item We propose a novel risk perspective exploration methods by scheduling risk levels. 
    \item We extensively evaluate the performance of scheduling risk levels in MARL environment and analyze the relation between risk levels and behaviors from it.
\end{itemize}




%% file: Manuscript/2_Related_Works.tex
\section{Related Works}
\label{sec:related_works}

\subsection{Distributional Reinforcement Learning}
\label{sec:distributional rl}
Distributional RL outputs a distribution of returns per an action that can be defined by a Dirac delta function as follows.
 \begin{equation}
    \begin{aligned}
    \label{dirac_delta}
     Z_{\theta }(x, a) := \sum_{i=1}^{N}p_{i}\delta_\theta {_{i(x, a)}}
    \end{aligned}
\end{equation}
where $\sum p_{i} = 1$. $\theta_{i}$,  $x$, and $a$ represent return value, a state, and an action, respectively. Generally, loss functions are defined as a Wasserstein distance which is given by,
 \begin{equation}
    \begin{aligned}
    \label{wasserstein}
     W_{p}(U, Y) = \left ( \int_{0}^{1} \left| F_{Y}^{-1}(\tau ) - F_{U}^{-1}(\tau )\right|^{p}d\tau  \right )^{1/p}
    \end{aligned}
\end{equation}
for TD-error of distribution with a Huber loss \citep{huber1992robust} where the inverse CDF $F_{Y}^{-1}$ of a random variable \textit{Y}  is defined as,
 \begin{equation}
    \begin{aligned}
    \label{inverseCDF}
     F_{Y}^{-1}(\tau ) := \textrm{inf}\left\{y \in \mathbb{R} : \tau \leq F_{Y}(y ) \right\}       
    \end{aligned}
\end{equation}
which is called a quantile function. Distributional RL is known to have low variance in performance, ensemble effect by making multiple outputs, and enabling risk-sensitive action selection as advantages.

\textbf{Risk-sensitiveness} is prevalent in economics and the stock market, where cautious or daring decisions are required. This approach was adopted in the domain of reinforcement learning known as risk-sensitive reinforcement learning for the benefit of selecting actions depending on risk. Generally, in the risk-sensitive RL area, risk-levels can be separated into 3 sections risk-averse, risk-neutral, and risk-seeking. Due to the variation in action space, a policy that is sensitive to risk can be read differently depending on the context. Nonetheless, we describe the general concept of risk-related policy in this section. A risk-averse policy can be interpreted as selecting the action with the highest state-action value among the worst-case scenarios per action. A risk-seeking policy entails selecting the same action as a risk-averse policy, but based on the best-case scenario. Risk-neutral policy positions amid risk-averse and seeking policy positions.

\textbf{Exploration in Distributional RL} \citet{zhou2021non} utilize Random Network Distillation \citep{burda2018exploration} algorithm to make two same architectures with randomly initialized parameters. They use the Wasserstein metric between two random networks' output per action to measure how many times the action was explored. \citet{mavrin2019distributional} use left truncated variance of the return distribution, which means optimistic action selection using outputs' distribution. 

\textbf{Distributional MARL}
Recently, some multi-agent reinforcement learning (MARL) algorithms adopted distribution-based architecture. \citet{sun2021dfac} integrated distributional RL and MARL by mean-shape decomposition. \citet{qiu2021rmix} adopt the conditional value at risk (CVaR) metric as a surrogate of joint action-state value $Q_{joint}$ and make a model that enables an adaptive risk level estimator with CVaR at every step.



\subsection{Environments} In a multi-agent system,  SMAC\citep{samvelyan2019starcraft} provides the most complex and dynamic environments for MARL researchers with partially observable MDP (POMDP), various scenarios, centralized training, and decentralized execution framework. Most MARL papers set SMAC as a default environment where they can prove their algorithm's performance. We experiment with our idea in this environment and show the relationship between risk levels and demanding skills.


%% file: Manuscript/3_Risk_scheduling.tex
\section{Risk perspective exploration}
\label{sec:risk_scheduling}


In this paper, we only have interest on the distributional reinforcement learning's perspective on exploration and risk-sensitiveness. With this risk concept, we try exploration for various risk levels by scheduling these risks accompanying exploration for less explored actions which show a similar effect of using variance.

\subsection{Selecting Risk levels}
In Distributional RL for a risk-sensitive algorithm, the output distribution is represented by a quantile function (inverse CDF) with a domain of $[0, 1]$, as stated in \autoref{sec:distributional rl}. Then, quantile fractions (which are typically expressed as $\tau$) can be sampled from $[0, 1]$. By sampling quantile fractions, we can induce risk-averse or risk-seeking behavior in the model. If we wish to make agents risk-averse, for instance, we can select quantile fractions from $[0, 0.25]$ which samples relatively low returns. Contrarily, sampling quantile fractions between $[0.75, 1]$ indicate risk-seeking behavior which induces sampling high returns.

\subsection{Exploration by risk scheduling}
For decades, value-based RL algorithms have depended on $\epsilon$-greedy\citep{watkins1989learning} decaying exploration or adding noise\citep{fortunato2017noisy} to the model parameters for exploration. Considering the method of decaying epsilon value from high to low, the thought flashes that with distributional RL, which can handle risk-sensitiveness, will it be possible to schedule or decay risk levels like the $\epsilon$-greedy exploration strategy?


In the $\epsilon$-greedy exploration of distributional RL, we can select actions as follows:
\begin{equation}
    \resizebox{0.90\columnwidth}{!}{$
    \begin{aligned}
    \label{epsilon}
        \begin{cases}
            \mathrm{arg\,max}_{a}\mathbb{E}_{\tau\sim \mathcal{U}[0,1]}[\mathbb{Z}(s, a)] & \mathrm{with\,\,probability} \,\, 1-\epsilon\\
            \mathrm{random\,\,action} & \mathrm{with\,\,probability} \,\, \epsilon\\
        \end{cases}   
    \end{aligned}
    $}
\end{equation}

where $s$ and $a$ mean action and state each, and $\mathbb{Z}(s, a)$ is the distribution of return given state and action, which makes action-state value $Q(s, a)$ by taking expectation. We select the action that makes state-action value $Q(s, a)$ the best with the probability $1-\epsilon$ and random action with the probability $\epsilon$. We adapt this decaying idea from $\epsilon$-greedy methods to risk scheduling which makes it to explore a variety of risk levels and also optimistic actions as follows:
\begin{equation}
    \resizebox{0.90\columnwidth}{!}{$
    \begin{aligned}
    \label{eqn:scheduling}
        \begin{cases}
            \mathrm{argmax}_{a}\mathbb{E}_{\tau\sim \mathcal{U}[\alpha,\beta]}[\mathbb{Z}(s, a)] & \mathrm{with\,\,probability} \,\, 1-\epsilon\\
            \mathrm{random\,\,action} & \mathrm{with\,\,probability} \,\, \epsilon\\
        \end{cases}   
    \end{aligned}
    $}
\end{equation}


where $\alpha$ and $\beta$ adjust the risk levels and keep changing through the scheduling steps. Like decaying epsilon from 1 to 0.05, decaying risks gradually from seeking level to specific risk level is the basic format of our method. When scheduling risks within the range of risk-seeking level, $\beta$ will change from 1 to 0 with $\alpha$ fixed to 0. $\beta$ will be set to 0 when scheduling the risk levels within the range of risk-averse level. The details of how to schedule risks are explained in \autoref{subsec:risk_scheduling}.




\begin{figure*}[t!]
    \centering
    \begin{subfigure}[h]{0.23\textwidth}
        \centering
        \includegraphics[width=\columnwidth, height = 2.5cm]{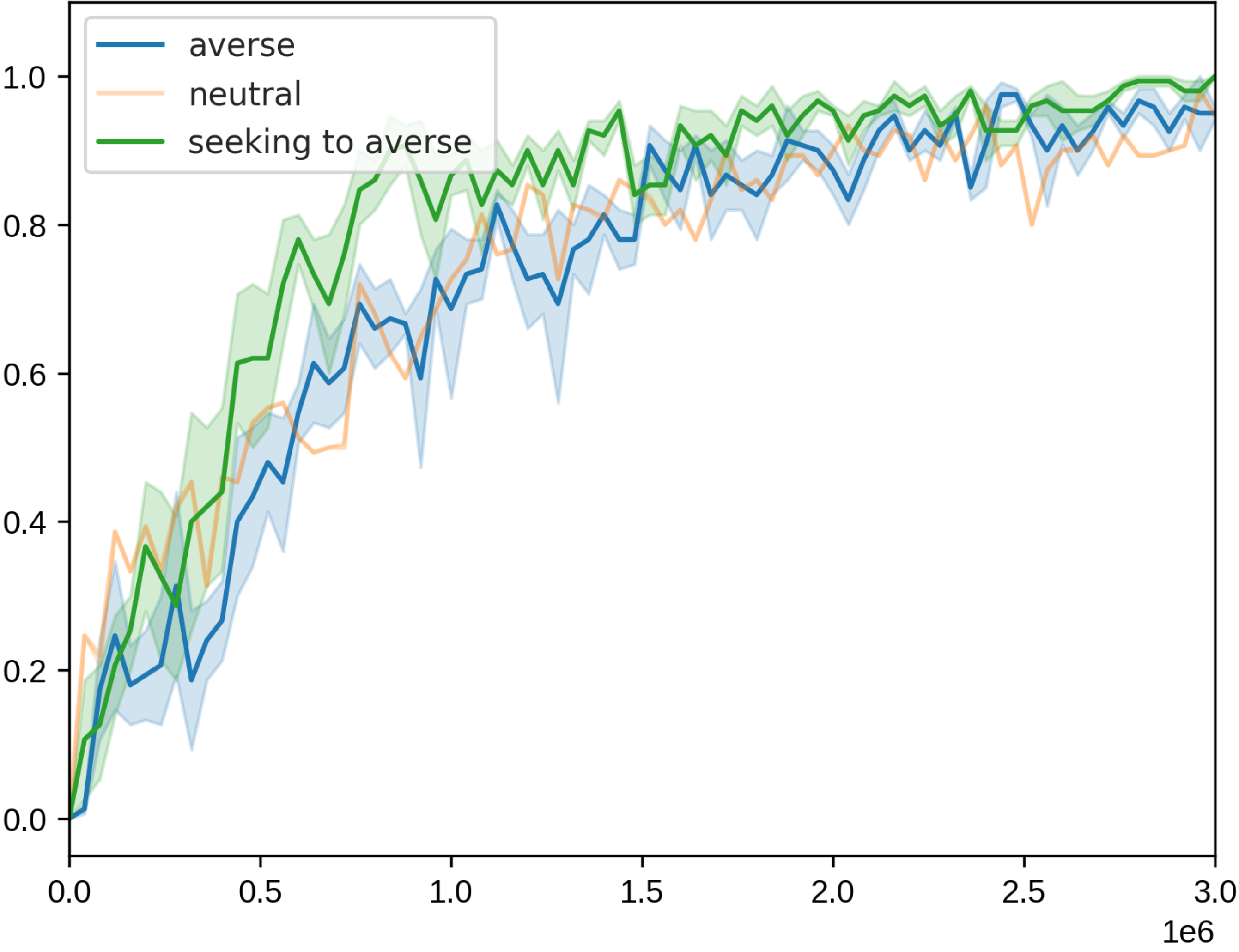}
        \vspace{-0.7cm}
        \caption{\texttt{2s3z} \scriptsize{(averse$^{5}$)}}
        \label{fig:2s3z_averse}
    \end{subfigure}
    \begin{subfigure}[h]{0.23\textwidth}
        \centering
        \includegraphics[width=\columnwidth, height = 2.5cm]{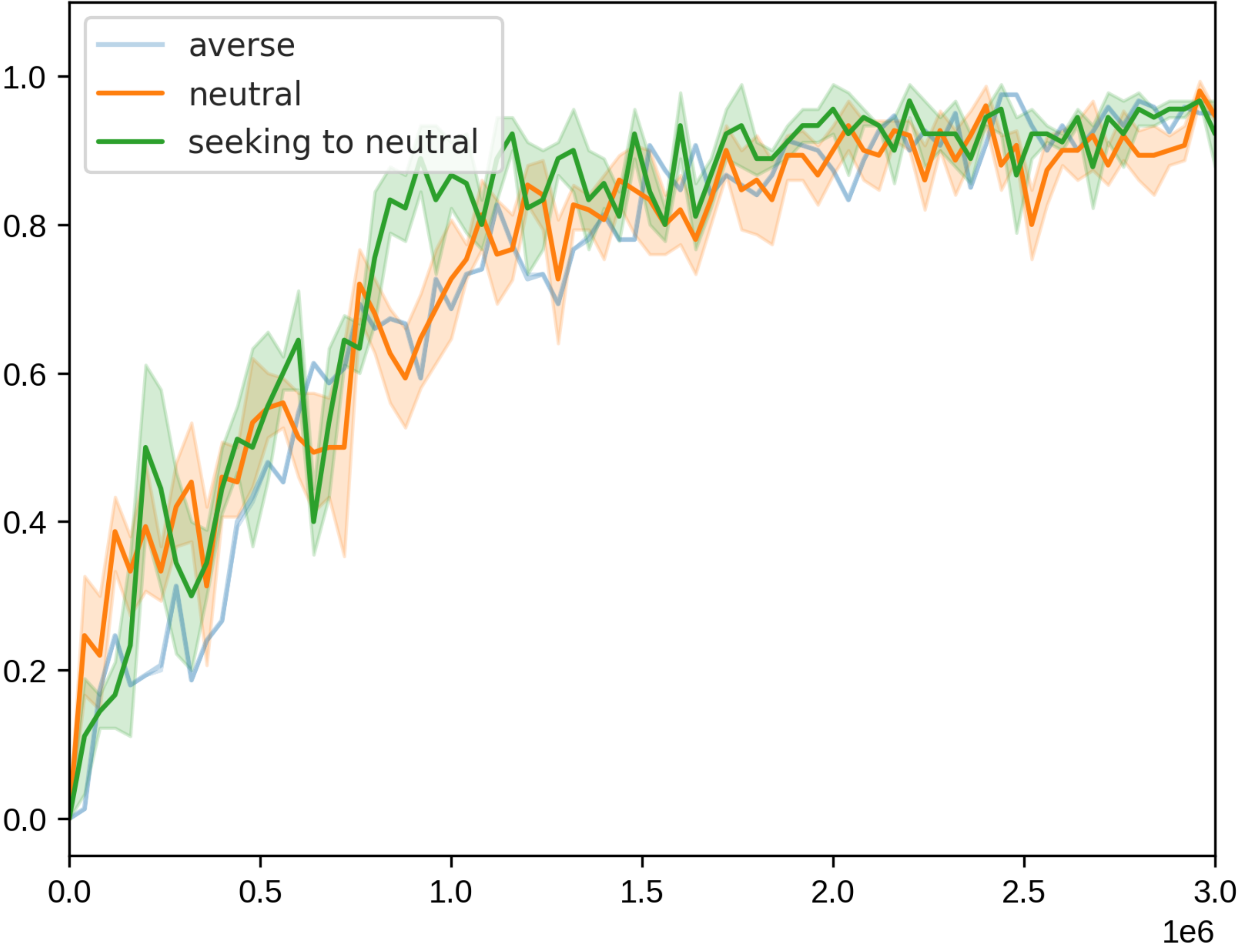}
        \vspace{-0.7cm}
        \caption{\texttt{2s3z} \scriptsize{(neutral$^{1}$)}}
        \label{fig:2s3z_neutral}
    \end{subfigure}
    \begin{subfigure}[h]{0.23\textwidth}
        \centering
        \includegraphics[width=\columnwidth, height = 2.5cm]{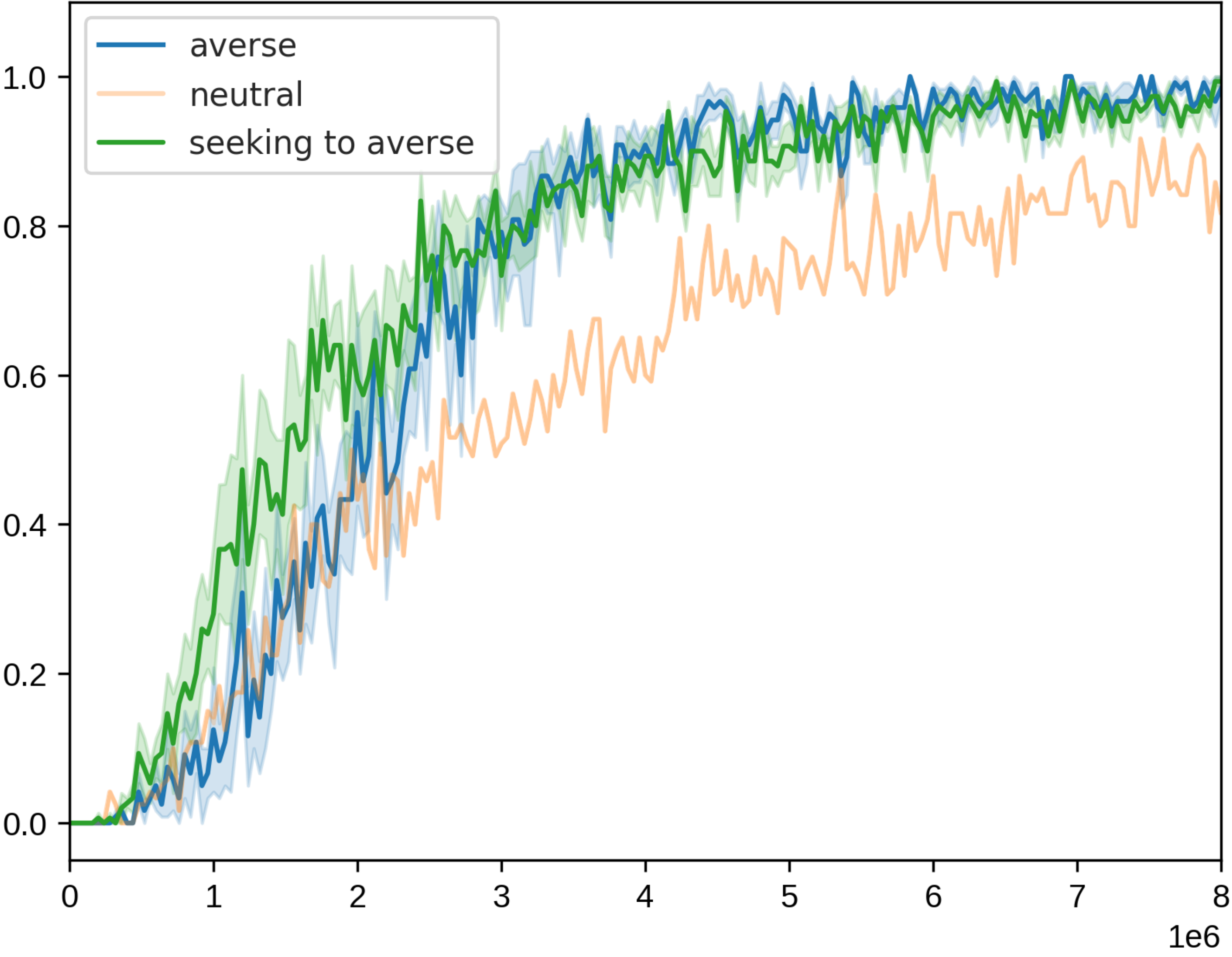}
        \vspace{-0.7cm}
        \caption{\texttt{3s5z} \scriptsize{(averse$^{5}$)}}
        \label{fig:3s5z_averse}
    \end{subfigure}
    \begin{subfigure}[h]{0.23\textwidth}
        \centering
        \includegraphics[width=\columnwidth, height = 2.5cm]{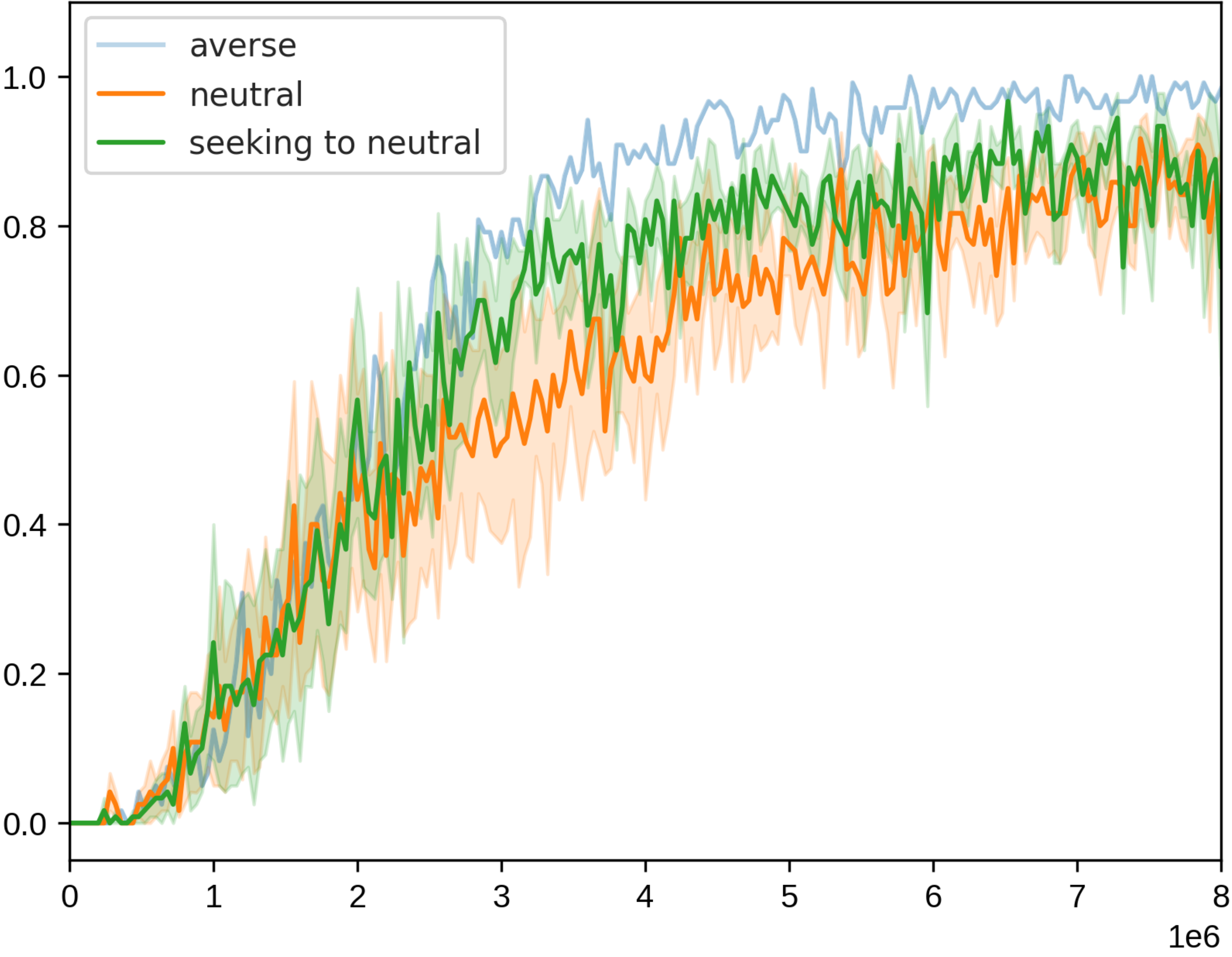}
        \vspace{-0.7cm}\\
        \caption{\texttt{3s5z} \scriptsize{(neutral$^{5}$)}}
        \label{fig:3s5z_neutral}
    \end{subfigure}
    \begin{subfigure}[h]{0.23\textwidth}
        \centering
        \includegraphics[width=\columnwidth, height = 2.5cm]{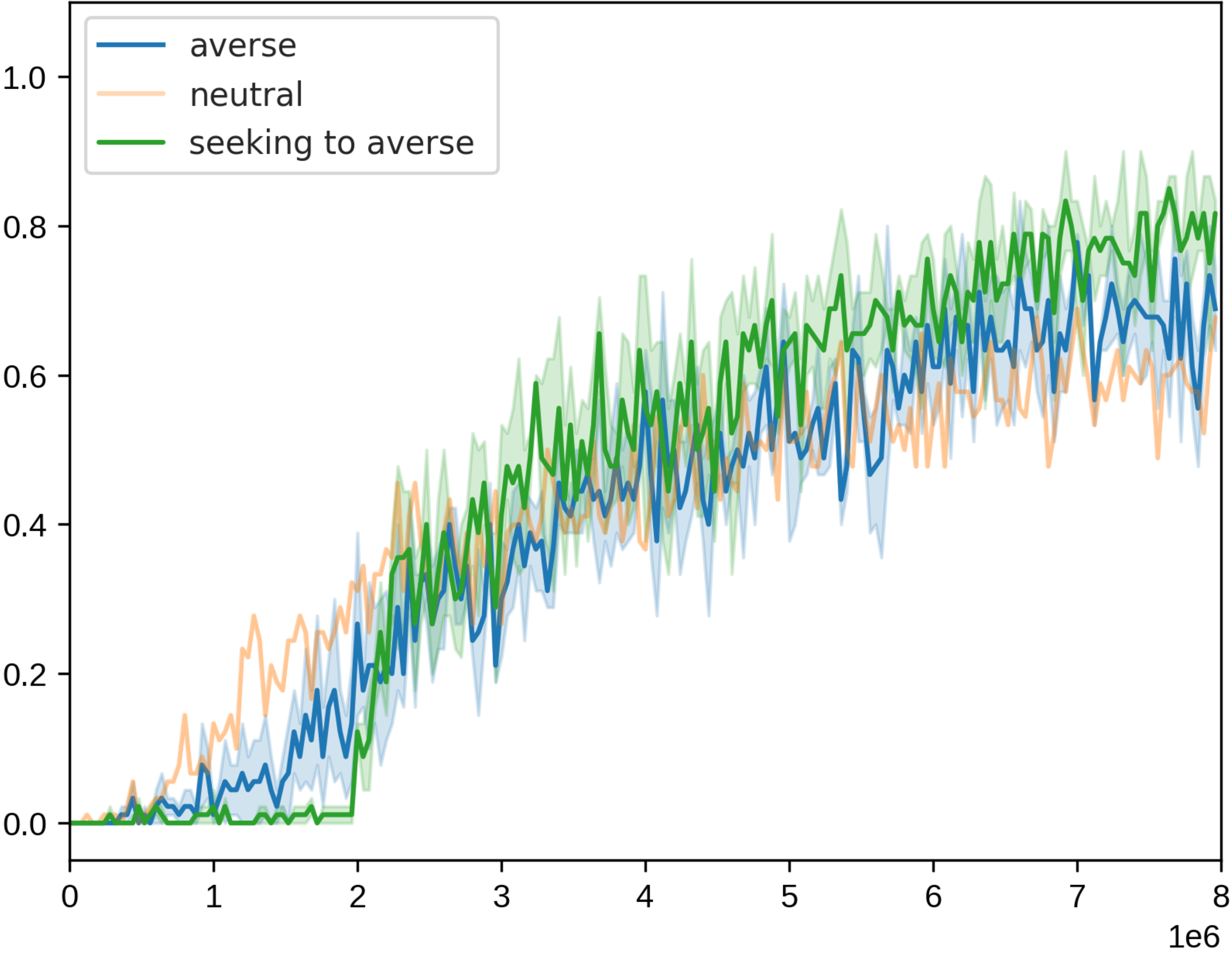}
        \vspace{-0.7cm}
        \caption{\texttt{5m vs 6m} \scriptsize{(averse$^{1}$)}}
        \label{fig:5mvs6m_averse}
    \end{subfigure}
    \begin{subfigure}[h]{0.23\textwidth}
        \centering
        \includegraphics[width=\columnwidth, height = 2.5cm]{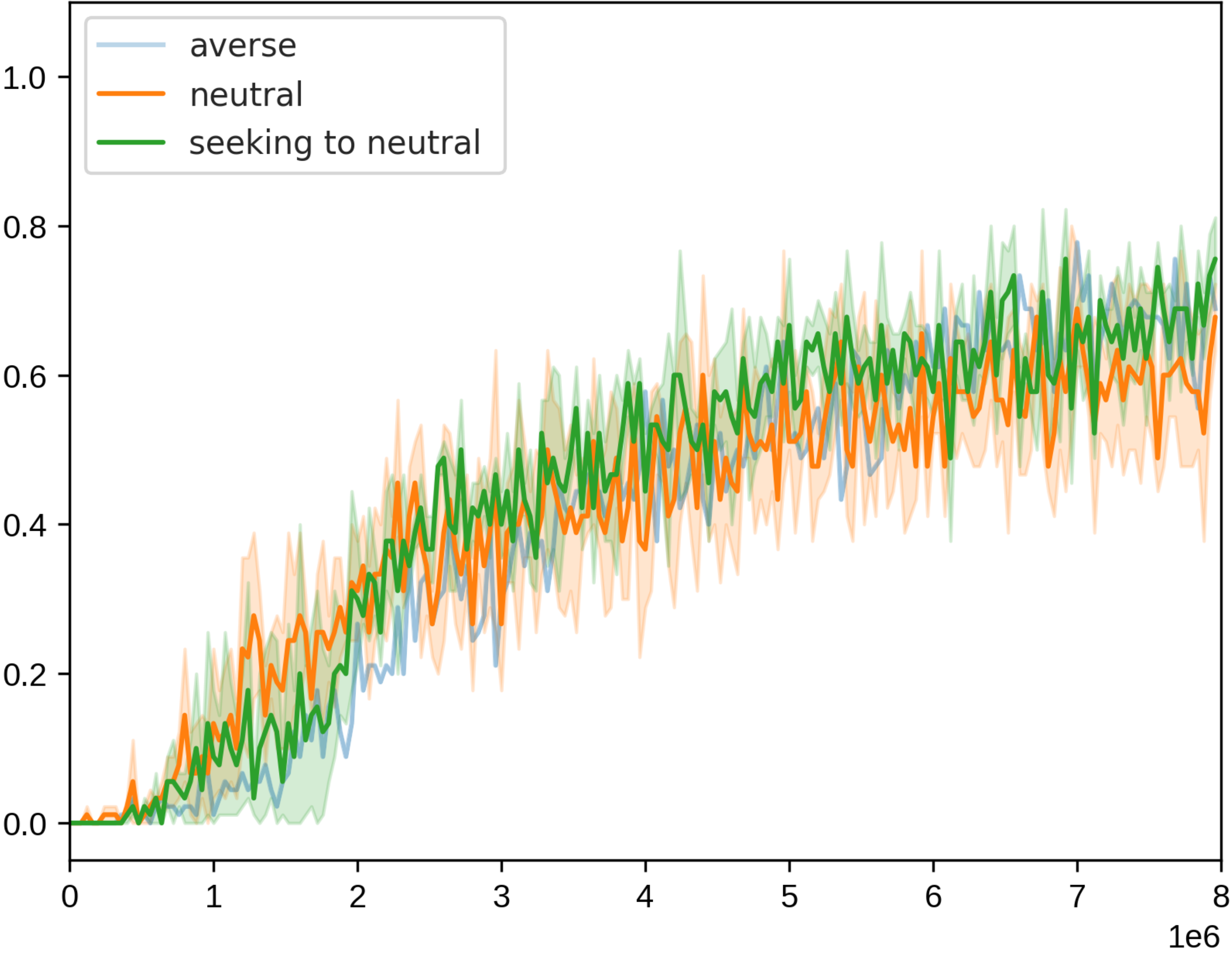}
        \vspace{-0.7cm}
        \caption{\texttt{5m vs 6m} \scriptsize{(neutral$^{1}$)}}
        \label{fig:5mvs6m_neutral}
    \end{subfigure}
    \begin{subfigure}[h]{0.23\textwidth}
        \centering
        \includegraphics[width=\columnwidth, height = 2.5cm]{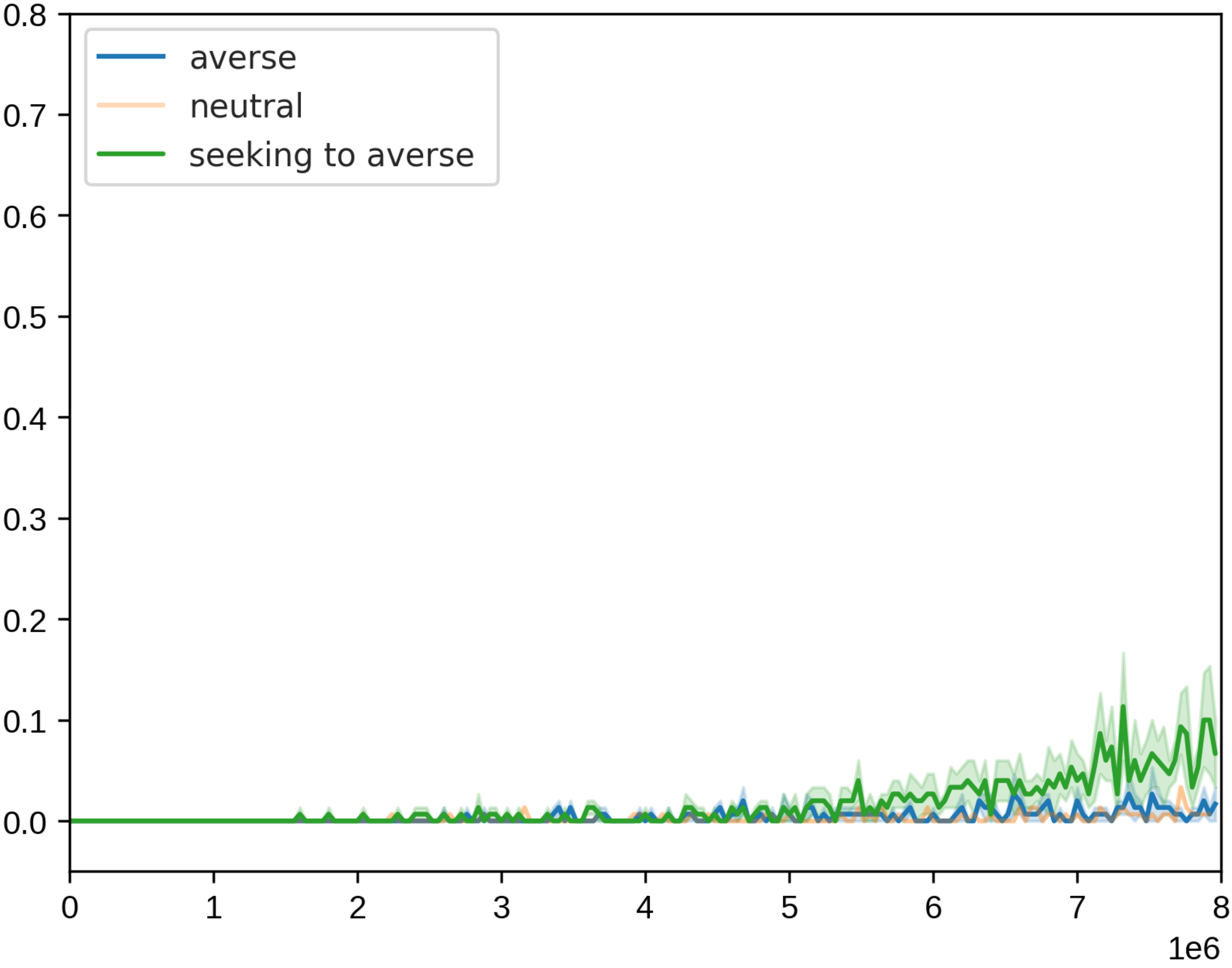}
        \vspace{-0.7cm}
        \caption{\texttt{3s5z vs 3s6z} \scriptsize{(averse$^{1}$)}}
        \label{fig:3s5zvs3s6z_averse}
    \end{subfigure}
    \begin{subfigure}[h]{0.23\textwidth}
        \centering
        \includegraphics[width=\columnwidth, height = 2.5cm]{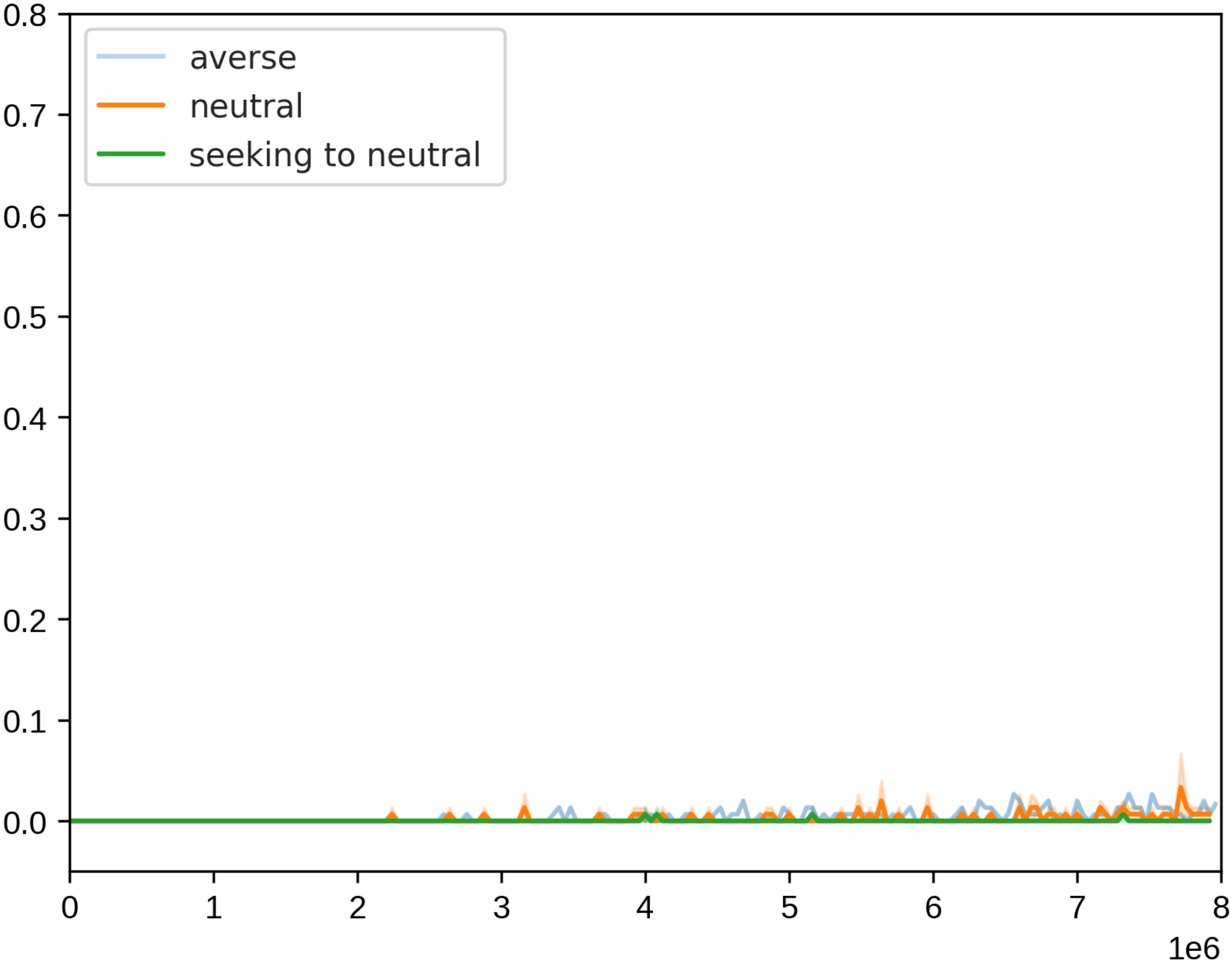}
        \vspace{-0.7cm}\\
        \caption{\texttt{3s5z vs 3s6z} \scriptsize{(neutral$^{1}$)}}
        \label{fig:3s5zvs3s6z_neutral}
    \end{subfigure}
    \begin{subfigure}[h]{0.23\textwidth}
        \centering
        \includegraphics[width=\columnwidth, height = 2.5cm]{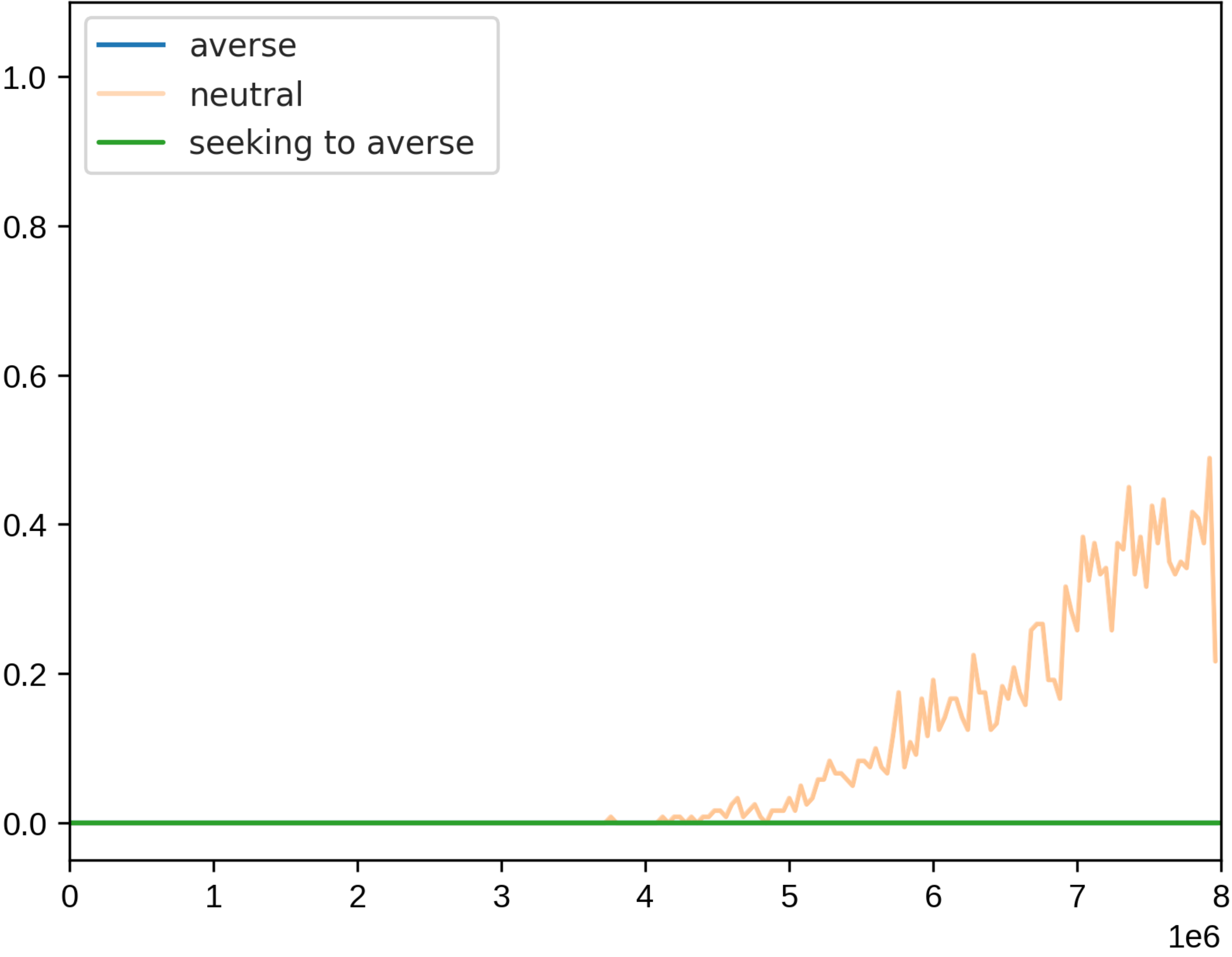}
        \vspace{-0.7cm}
        \caption{\texttt{3s vs 5z} \scriptsize{(averse$^{1}$)}}
        \label{fig:3svs5z_averse}
    \end{subfigure}
    \begin{subfigure}[h]{0.23\textwidth}
        \centering
        \includegraphics[width=\columnwidth, height = 2.5cm]{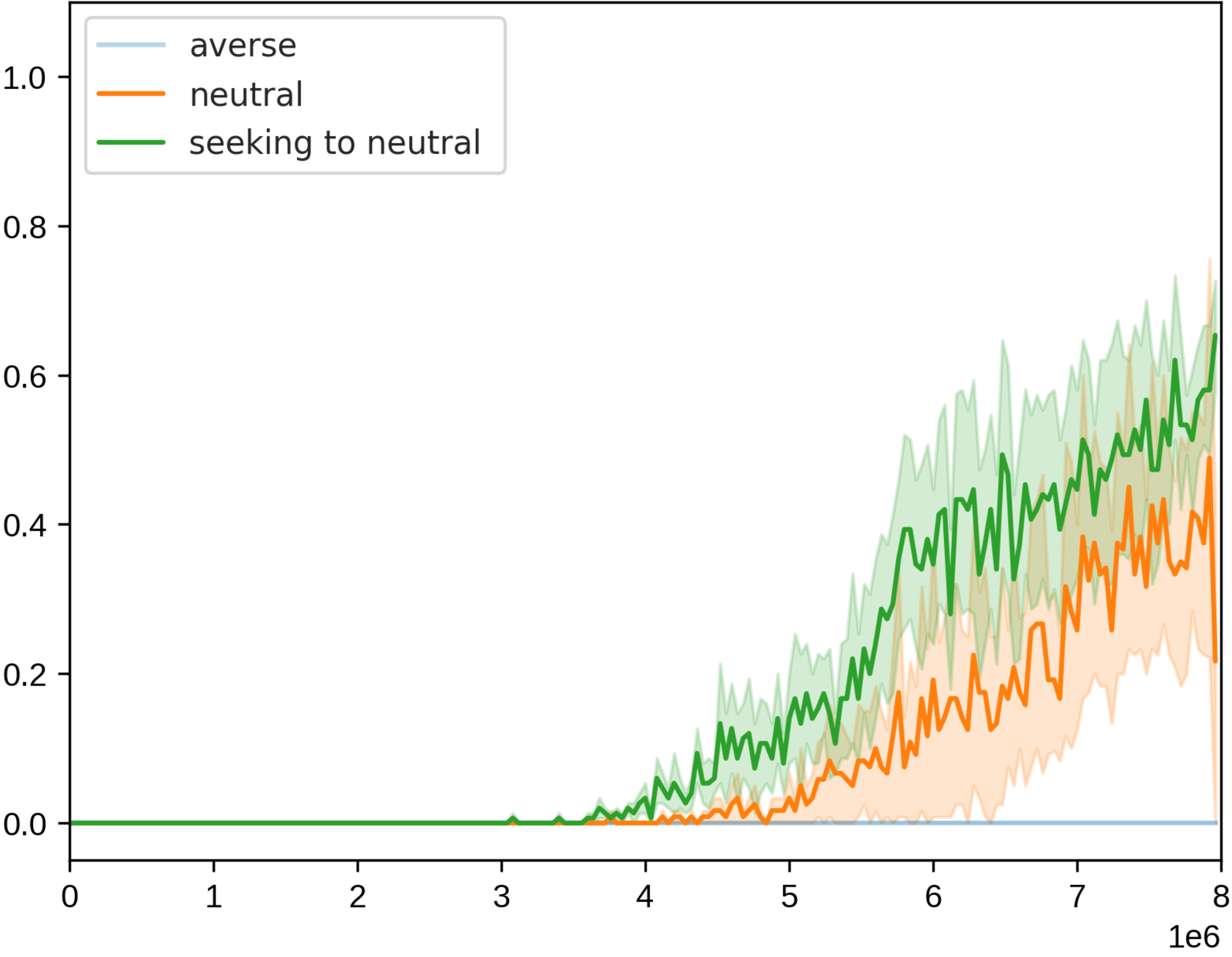}
        \vspace{-0.7cm}
        \caption{\texttt{3s vs 5z} \scriptsize{(neutral$^{1}$)}}
        \label{fig:3svs5z_averse}
    \end{subfigure}
    \begin{subfigure}[h]{0.23\textwidth}
        \centering
        \includegraphics[width=\columnwidth, height = 2.5cm]{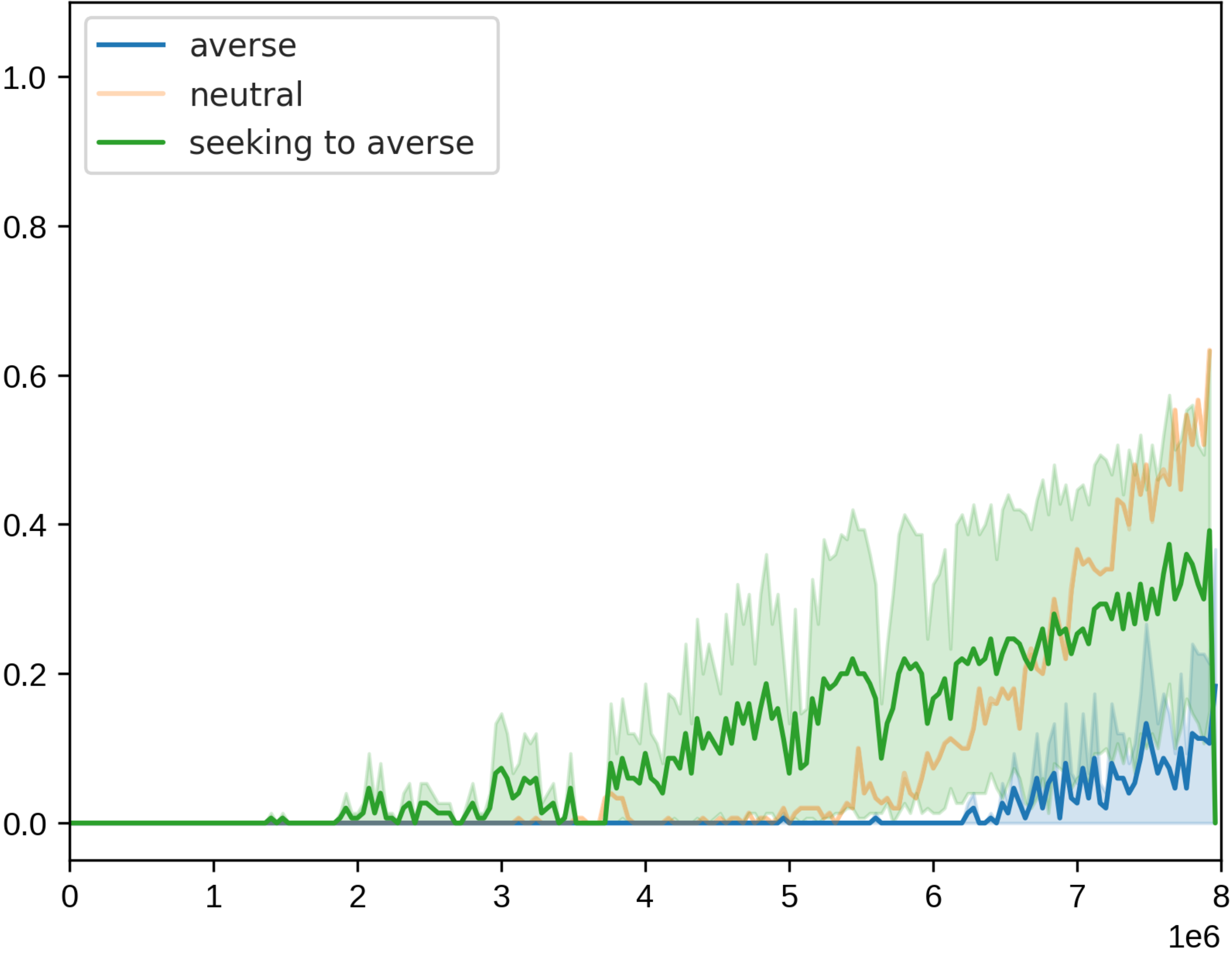}
        \vspace{-0.7cm}
        \caption{\texttt{corridor} \scriptsize{(averse$^{5}$)}}
        \label{fig:corridor_averse}
    \end{subfigure}
    \begin{subfigure}[h]{0.23\textwidth}
        \centering
        \includegraphics[width=\columnwidth, height = 2.5cm]{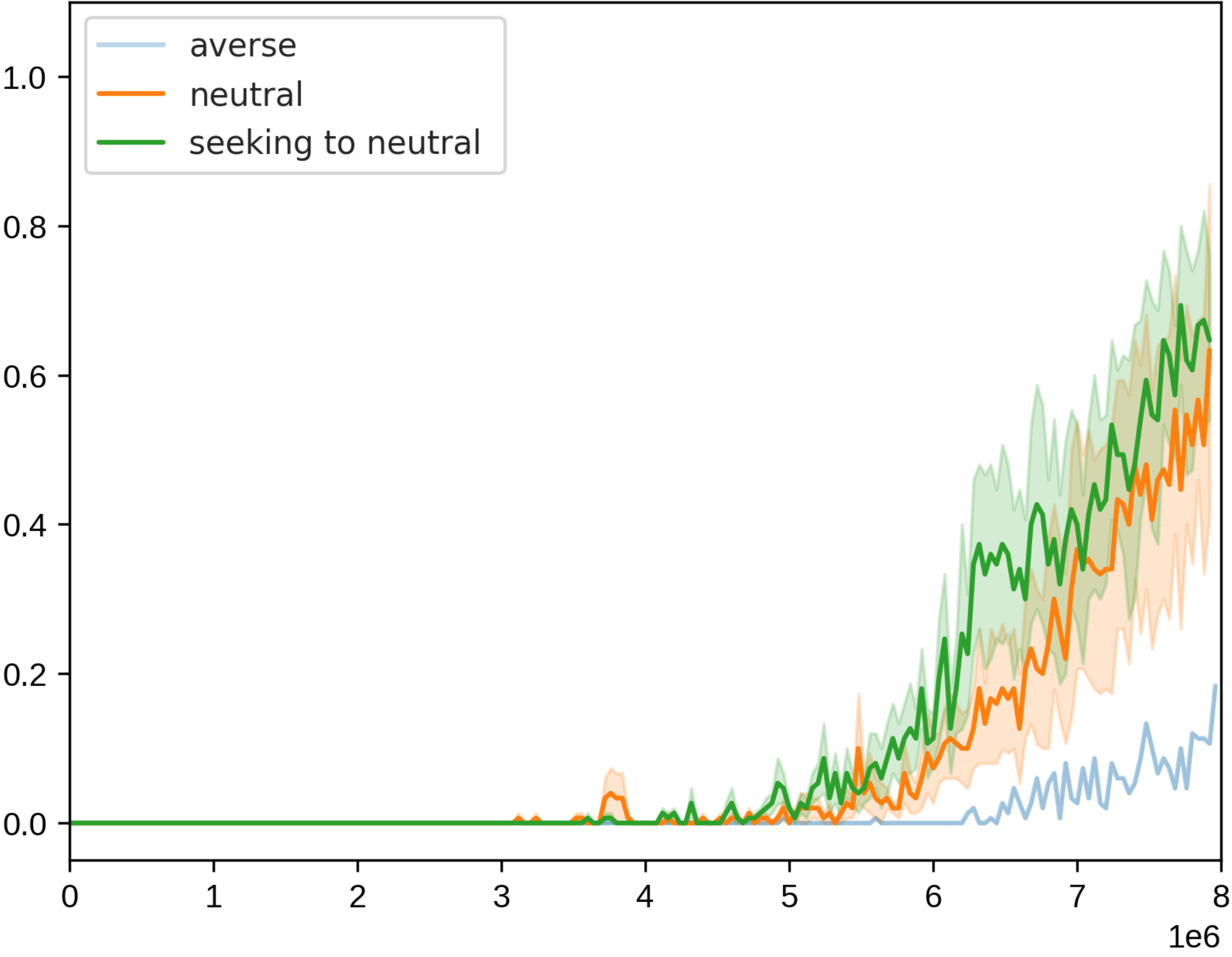}
        \vspace{-0.7cm}\\
        \caption{\texttt{corridor} \scriptsize{(neutral$^{5}$)}}
        \label{fig:corridor_neutral}
    \end{subfigure}
    \begin{subfigure}[h]{0.23\textwidth}
        \centering
        \includegraphics[width=\columnwidth, height = 2.5cm]{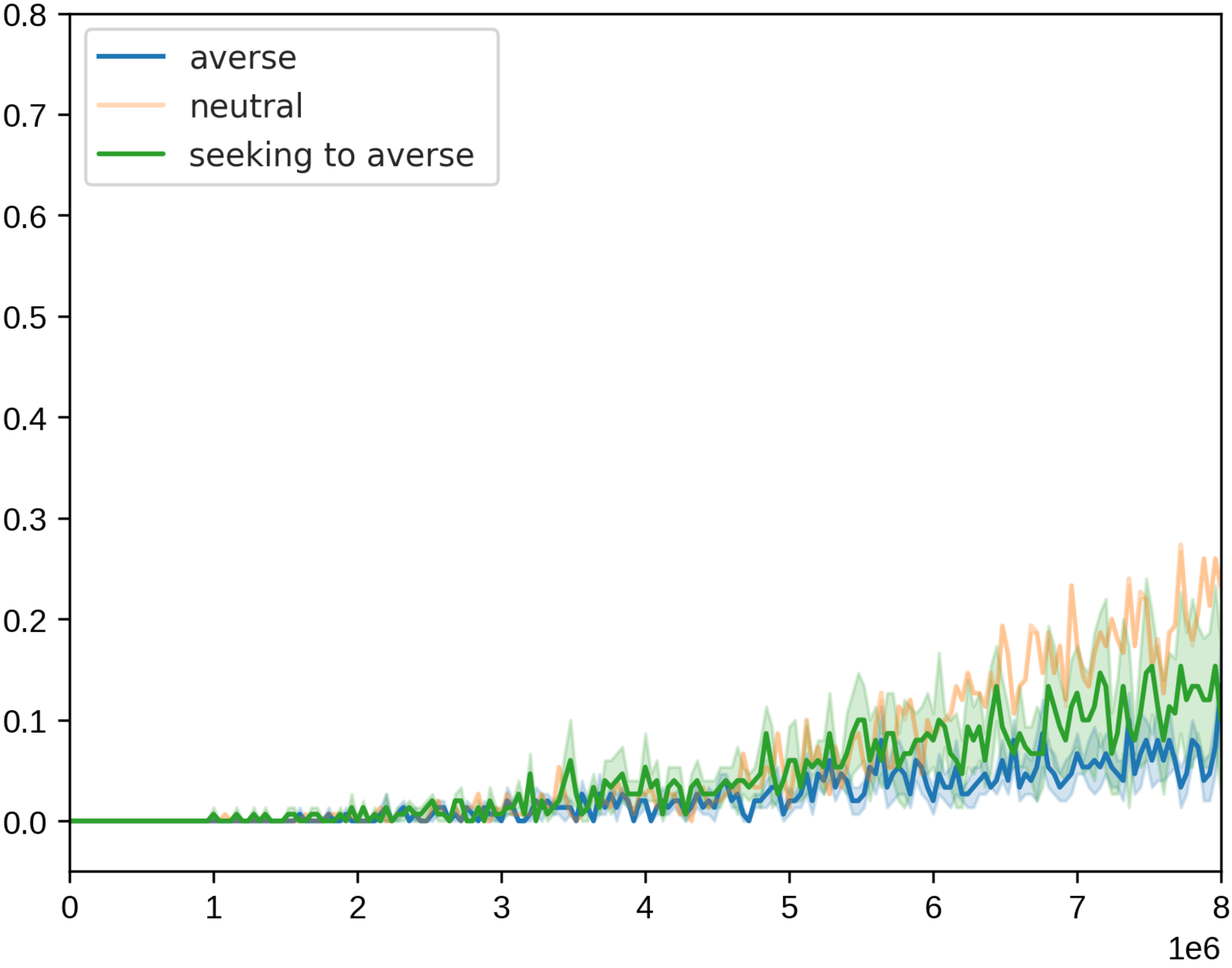}
        \vspace{-0.7cm}
        \caption{\texttt{6h vs 8z} \scriptsize{(averse$^{5}$)}}
        \label{fig:6hvs8z_averse}
    \end{subfigure}
    \begin{subfigure}[h]{0.23\textwidth}
        \centering
        \includegraphics[width=\columnwidth, height = 2.5cm]{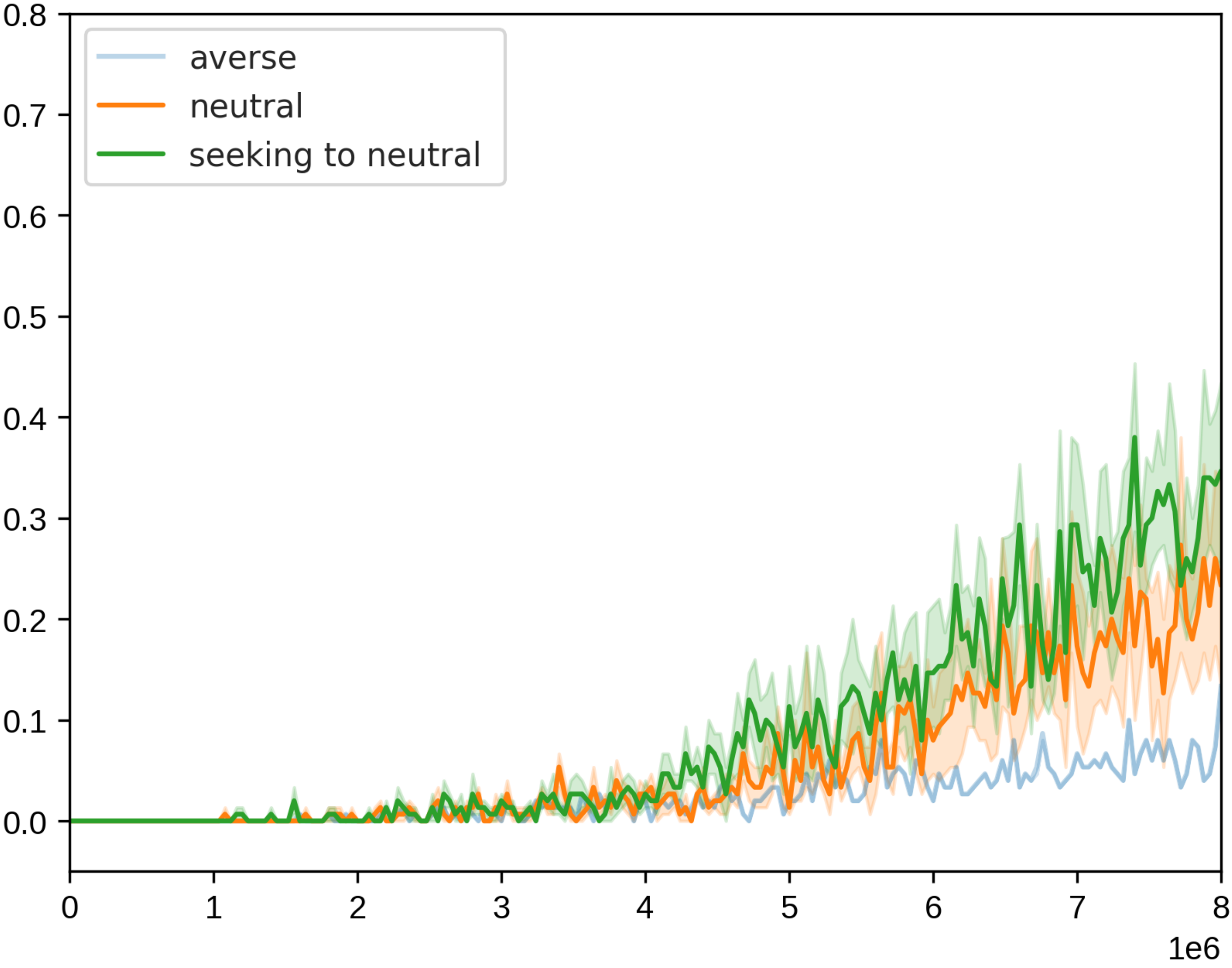}
        \vspace{-0.7cm}
        \caption{\texttt{6h vs 8z} \scriptsize{(neutral$^{5}$)}}
        \label{fig:6hvs8z_neutral}
    \end{subfigure}
    \begin{subfigure}[h]{0.23\textwidth}
        \centering
        \includegraphics[width=\columnwidth, height = 2.5cm]{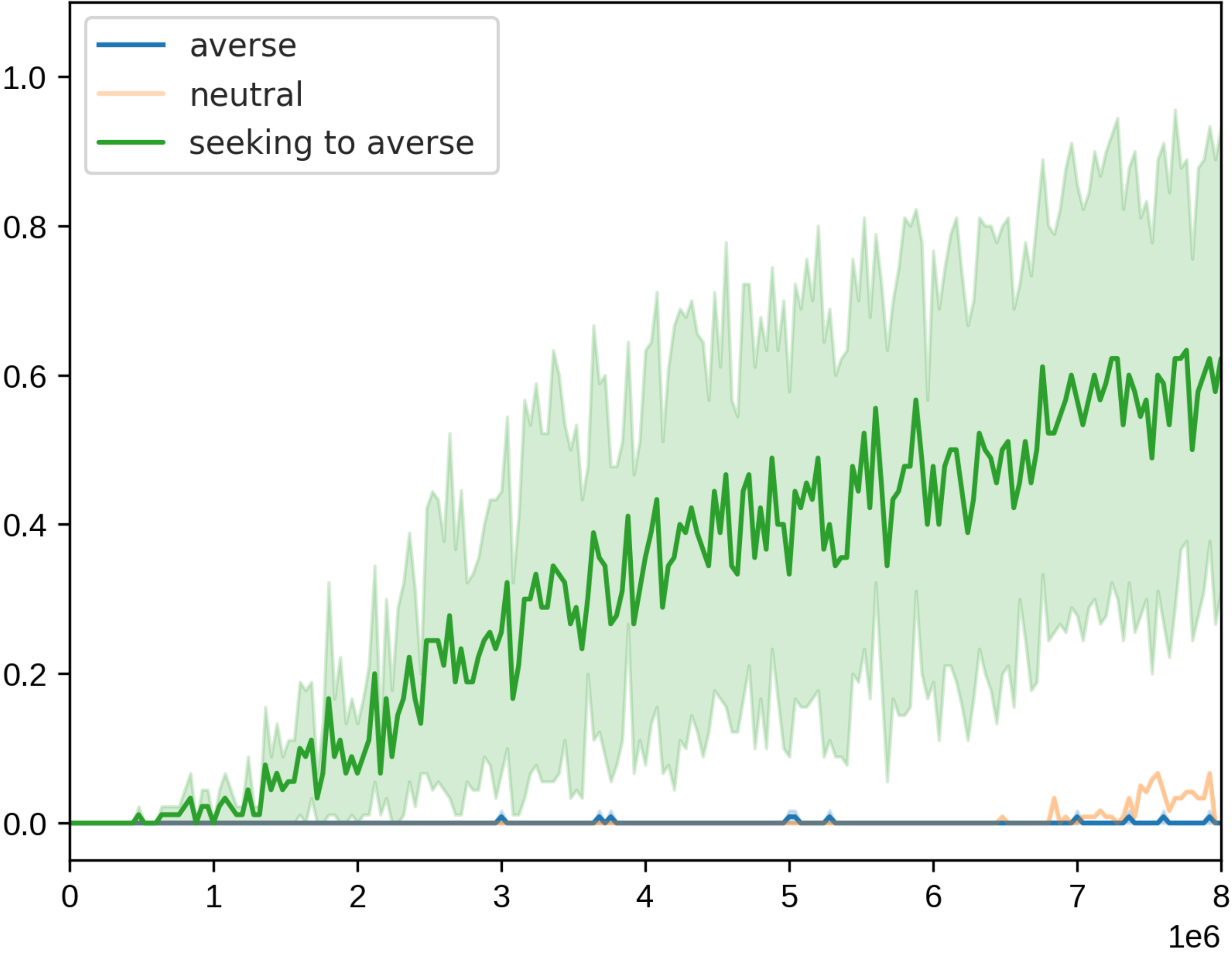}
        \vspace{-0.7cm}
        \caption{\texttt{MMM2} \scriptsize{(averse$^{1}$)}}
        \label{fig:MMM2_averse}
    \end{subfigure}
    \begin{subfigure}[h]{0.23\textwidth}
        \centering
        \includegraphics[width=\columnwidth, height = 2.5cm]{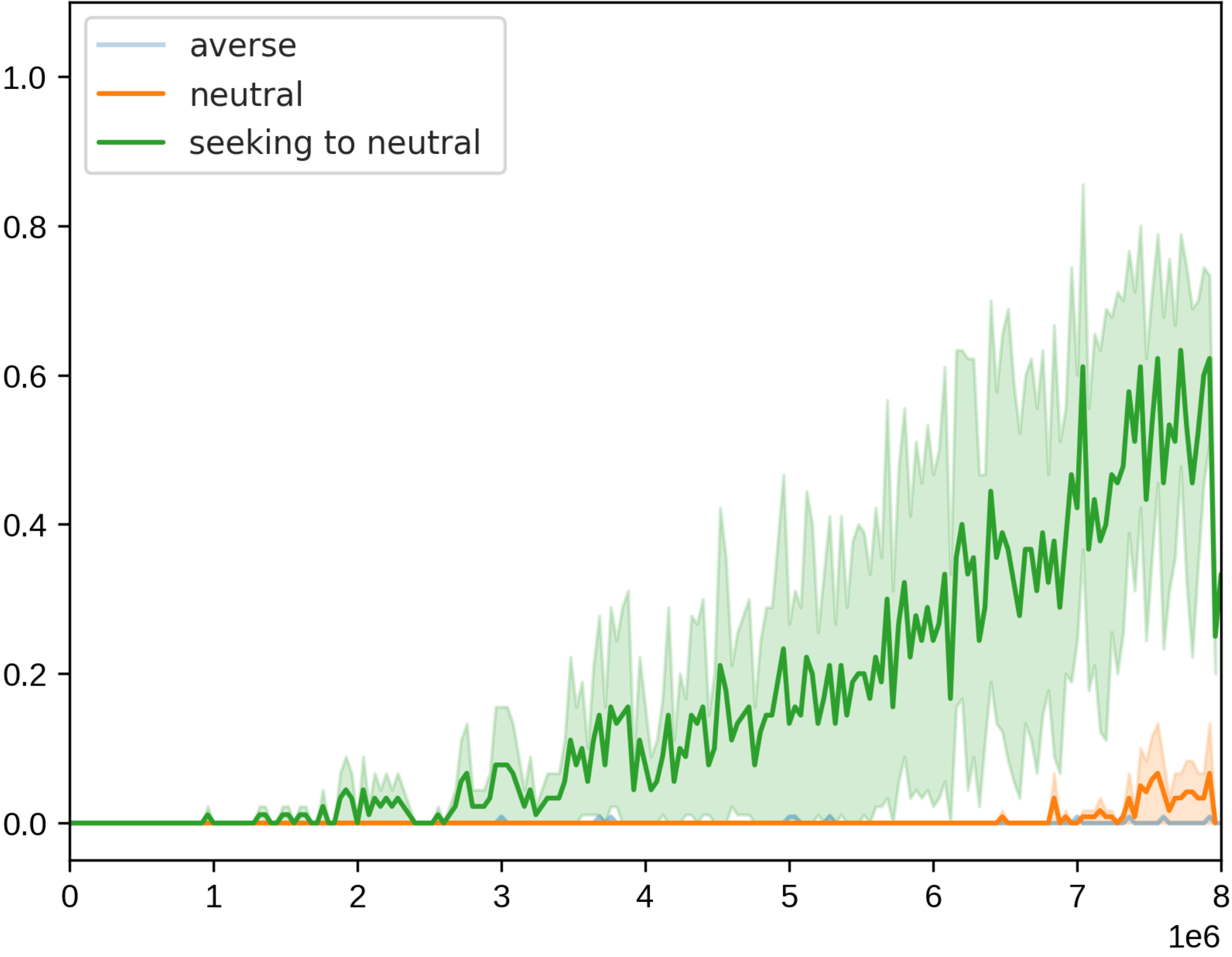}
        \vspace{-0.7cm}
        \caption{\texttt{MMM2} \scriptsize{(neutral$^{1}$)}}
        \label{fig:MMM2_neutral}
    \end{subfigure}
    \caption{A Win-rate graph where the x-axis is training steps. We compare the averse and neutral policy with scheduling strategy from risk-seeking to averse or neutral. Lines are made by means of 5 runs, and shaded areas show a 75\% confidence interval using five parallel training. We set the maximum win-rate to 0.8 for the scenarios that show performance lower than 0.5;otherwise, set it to 1.0. The number 1 and 5 located on the upper right of the letter `averse' and `neutral' means 10k and 50k scheduling timestep each.} 
    \label{fig:results}
\end{figure*}

%% file: Manuscript/4_Experiments.tex
\section{Experiments}
\label{sec:experiments}

We evaluate our method with the MARL algorithm, which suffers from intricate exploration caused by a cooperative goal, multiple agents, and POMDP setting. In this experiment, the DMIX algorithm, a DFAC variant, is used. We perform experiments in SMAC \cite{samvelyan2019starcraft} environment, with \texttt{2s3z, 3s5z} (\textit{easy}), \texttt{5m vs 6m, 3s vs 5z} (\textit{hard}), and \texttt{corridor, 3s5z vs 3s6z, 6h vs 8z, MMM2} (\textit{super hard}) scenarios each. Except for running SMAC environment in a parallel manner with five runners, we follow the default setting of DFAC with 8 million training time step. 
\subsection{Risk Scheduling}
\label{subsec:risk_scheduling}
We set \textbf{risk-averse}, \textbf{risk-neutral}, and \textbf{risk-seeking} as a way of sampling quantile fractions ($\tau$) from a uniform distribution $\mathcal{U}[0, 0.25]$, $\mathcal{U}[0, 1]$, and $\mathcal{U}[0.75, 1]$ each. Although there are many decaying skills in scheduling, we select the most basic decaying method, \textit{Linear} decaying. For example, if we want to schedule risk from seeking to averse, we set $\alpha$ and $\beta$ in \autoref{eqn:scheduling} to 0.75 and 1.0 each. Then sample quantile fractions from $[0.75, 1]$ and $\alpha$ start decaying to $0$ linearly. When $\alpha$ becomes 0, $\beta$ begins to move toward 0.25 linearly. The sampling range reaches $[0, 0.25]$ at the final scheduling step and is fixed until the training is finished. Seeking to averse and seeking to neutral risk policies are adopted in the experiments to compare with static risk level policies. Because the risk-seeking policy does not work in any scenarios, we did not insert the results of risk-seeking policy in this paper. We search decaying steps in a set of \{10k, 25k, 50k\}. 
    
    

\subsection{Results}
\label{main results}
\autoref{fig:results} demonstrates the learning curves in SMAC scenarios. Figures from \ref{fig:2s3z_averse} to \ref{fig:3s5zvs3s6z_neutral} show better performance when taking risk-averse policy, and Figures from \ref{fig:3svs5z_averse} to \ref{fig:MMM2_neutral} show better performance when taking risk-neutral policy. Also, it can be observed that the learning curves with risk scheduling grow faster and have higher final win-rates than static policies, as shown in \autoref{table:win-rate}. In addition, we find that there is a tendency that the more complex the scenario is, (the less likely the winning rate is), the performance improvements increase.

\begin{table}[!t]
    \centering
    \small
    \caption{Final win-rate performance}
    \label{table:win-rate}
    \resizebox{1\columnwidth}{!}{

    \vspace{5pt}
    \addtolength{\tabcolsep}{4pt}
    \resizebox{\linewidth}{!}{
    \begin{tabular}{lccccc}
    \toprule
    Maps & neutral & neutral$^{*}$ & averse & averse$^{*}$ \\
    \midrule
    \texttt{2s3z} & 0.951\ & 0.948 & 0.951 & \textbf{0.988}\\
    \texttt{3s5z} & 0.822 & 0.849 & 0.977 & \textbf{0.982}\\
    \texttt{5m vs 6m} & 0.607 & 0.718 & 0.696 & \textbf{0.792}\\
    \texttt{3s vs 5z} & 0.422 & \textbf{0.604} & 0.0 & 0.0\\
    \texttt{corridor} & 0.533 & \textbf{0.662} & 0.122 & 0.320\\
    \texttt{6h vs 8z} & 0.235 & \textbf{0.340} & 0.084 & 0.124\\
    \texttt{MMM2} & 0.044 & 0.518 & 0.002 & \textbf{0.607}\\
    \texttt{3s5z vs 3s6z} & 0.006 & 0.0 & 0.013 & \textbf{0.086}\\
    \bottomrule
    \multicolumn{4}{l}{\footnotesize $*$ : scheduling method} \\

    \end{tabular}}}
\end{table}

\subsection{Analysis}
\begin{figure}[H]
    \centering
    \begin{subfigure}[h]{\columnwidth}
        \centering
        \includegraphics[width=0.47\columnwidth]{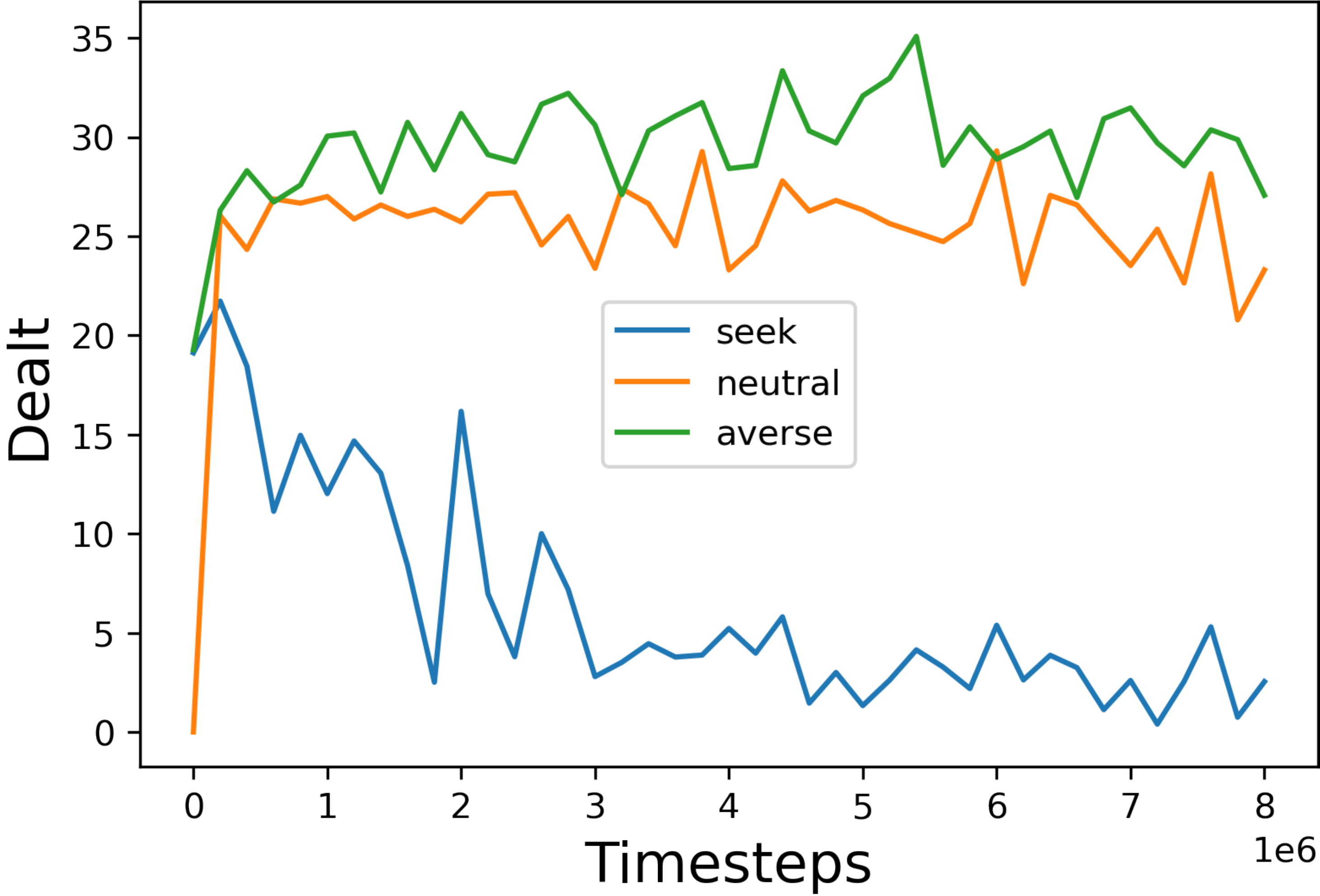}
        \centering
        \includegraphics[width=0.47\columnwidth]{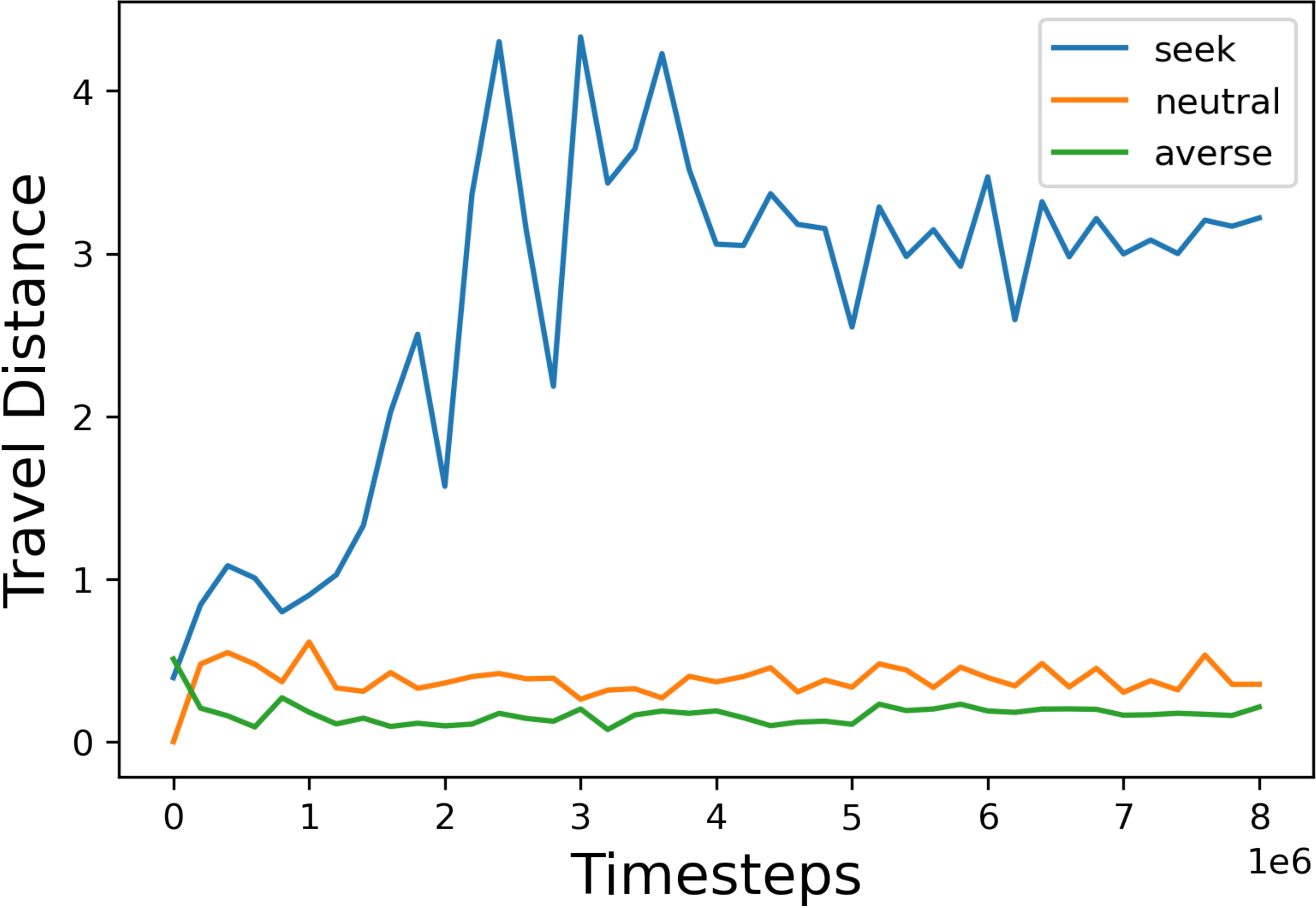}
        \caption{}
        \label{fig:learning}
    \end{subfigure}
    \begin{subfigure}[h]{\columnwidth}
        \centering
        \includegraphics[width=0.47\columnwidth]{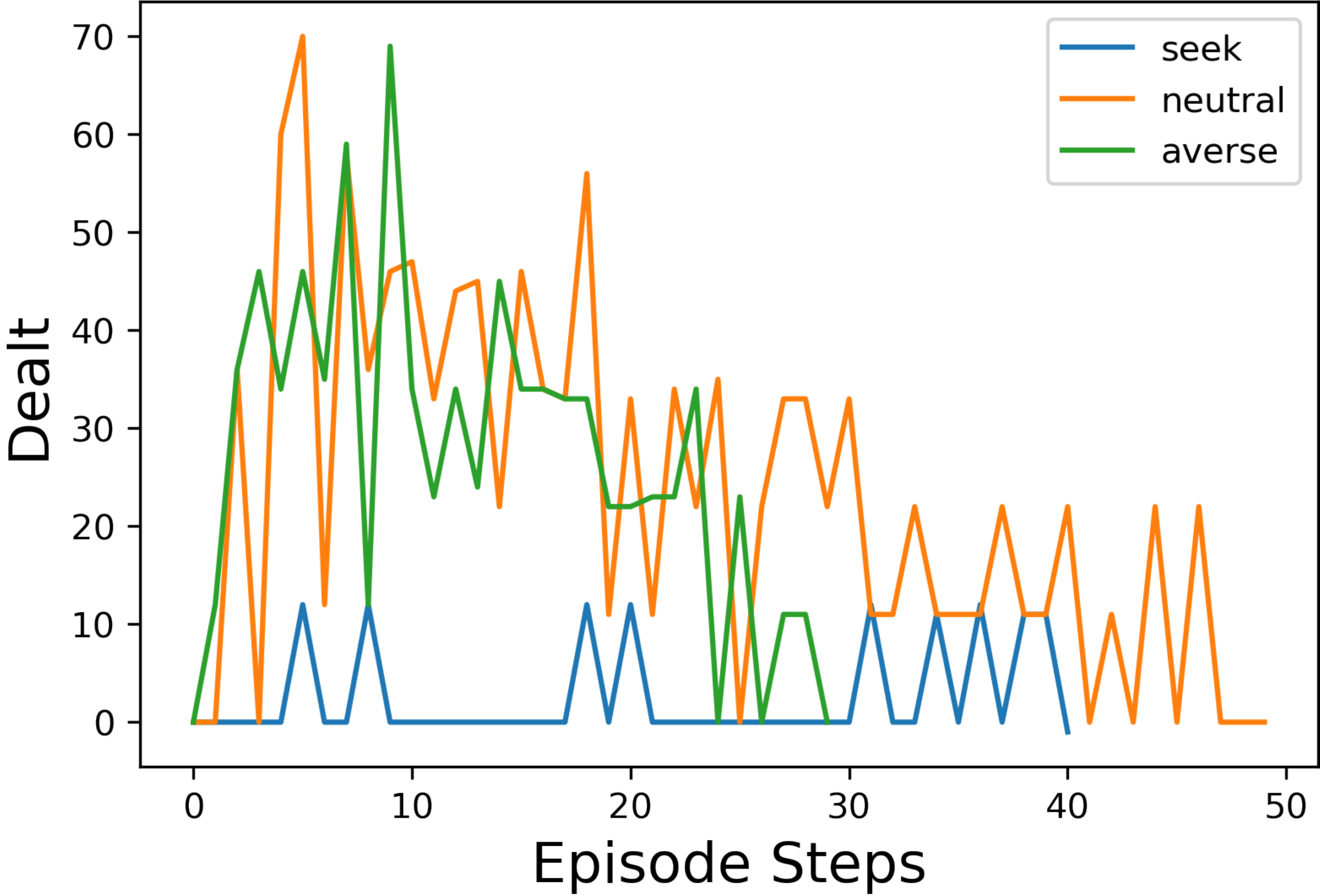}
        \centering
        \includegraphics[width=0.47\columnwidth]{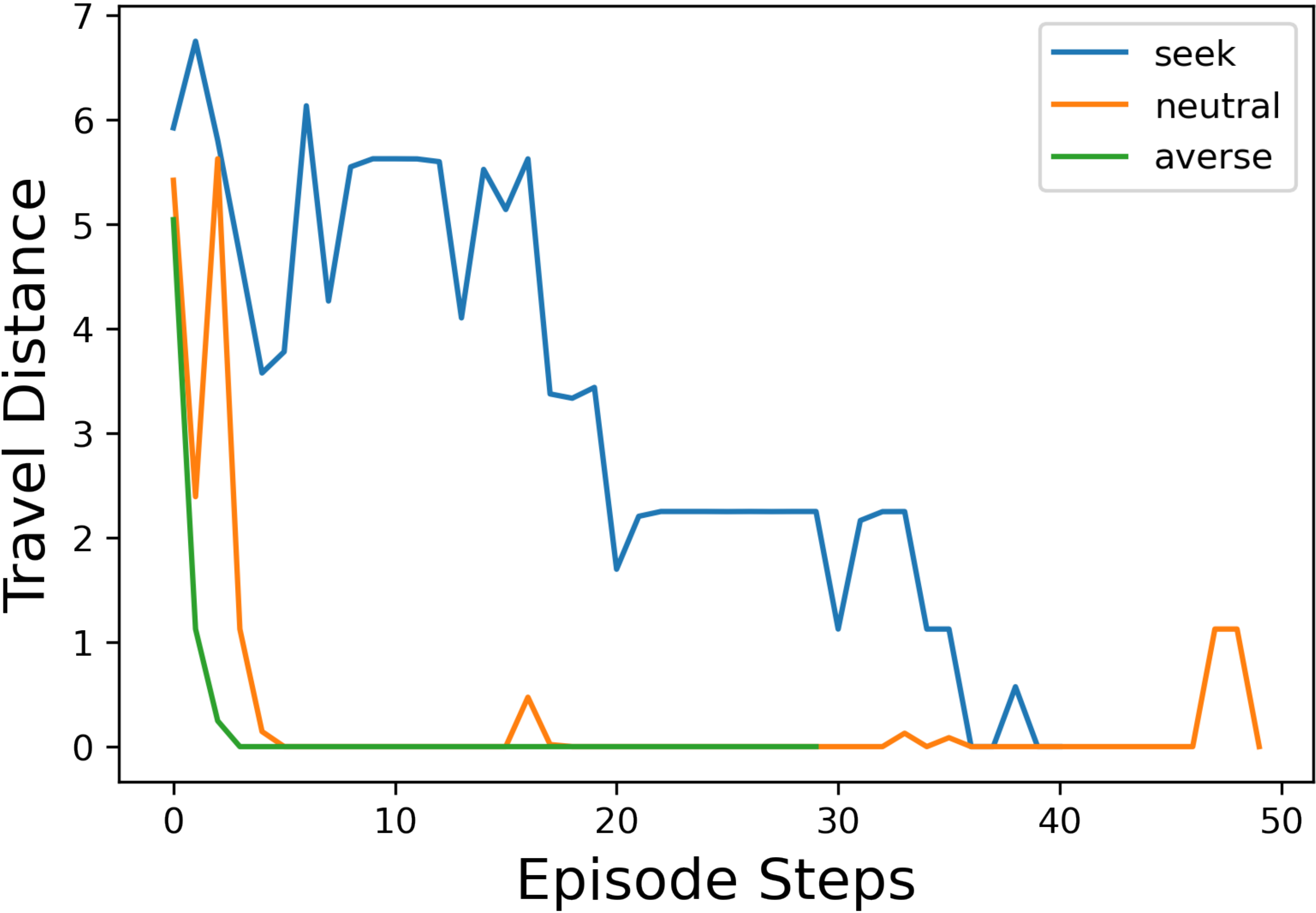}
        \caption{}
        \label{fig:episode}
    \end{subfigure}
    \caption{Results on \texttt{6h vs 8z} scenario. (a) left shows average dealt from agents to enemies and right shows travel distance of agents per timestep in a single episode each. (b) left shows total dealt from agents to enemies and right shows travel distance of agents at each timestep in a single episode.}
    \label{fig:dealt_moving_results}
    \vspace{-0.5cm}
\end{figure}

\paragraph{Relation between risk level \& behavior}
\label{relation risk behavior}
We have evidence that risk levels are related to agents' behavior tendencies in the SMAC scenario. Risk-averse behavior tends to attack enemies more than other policies, as shown in \autoref{fig:learning} deploy intensive fire per timestep. This entirely corresponds to the fact that risk-averse policy shows the behavior that chooses the best action among the worst case, which can be just attacking the enemies. Seeking behavior tends to do other than attacking (e.g., Move around) relative to other policies seeing the \autoref{fig:learning}, which can be interpreted as a case that agents select action among the best case. Because if agents survive longer by the moving behavior, the possibility of receiving a good return could be higher than other actions. However, moving continuously by risk-seeking policy can result in a bad performance, which makes no immediate reward. Risk-neutral behavior shows intermediate temper that demonstrates adequate firepower per timestep and more moving distance than averse policy, as shown in \autoref{fig:learning}. \autoref{fig:episode}, conducted in a single episode, supports the above argument in detail. So \autoref{fig:dealt_moving_results} demonstrates why the tendency of \autoref{fig:results} comes out. With this evidence, we conclude that risk-averse and neutral policy is related to \textit{focusing fire} and \textit{kiting} skills each. Scenarios and challenging skills are shown in \autoref{table:skills and risk level}.

\begin{table}[!t]
    \centering
    \caption{Challenging skills \& risk level}
    \label{table:skills and risk level}
    \resizebox{0.85\columnwidth}{!}{

    \vspace{5pt}
    \addtolength{\tabcolsep}{4pt}
    \resizebox{\linewidth}{!}{
    \begin{tabular}{lcc}
    \toprule
    Maps & challenging skills & risk levels \\
    \midrule
    \texttt{2s3z} & focusing fire & averse $=$ neutral\\
    \texttt{3s5z} & focusing fire & averse \\
    \texttt{5m vs 6m} & focusing fire & averse \\
    \texttt{3s vs 5z} & kiting & neutral\\
    \texttt{corridor} & breaking each & neutral\\
    \texttt{6h vs 8z} & kiting & neutral\\
    \texttt{MMM2} & focusing fire & neutral\\
    \texttt{3s5z vs 3s6z} & focusing fire & averse\\
    \bottomrule

    \end{tabular}}}
\end{table}

\paragraph{Scheduling risk levels}
\begin{wrapfigure}{r}{0.5\columnwidth}
  \begin{center}
    \vspace{-0.20cm}
    \includegraphics[width=0.5\columnwidth]{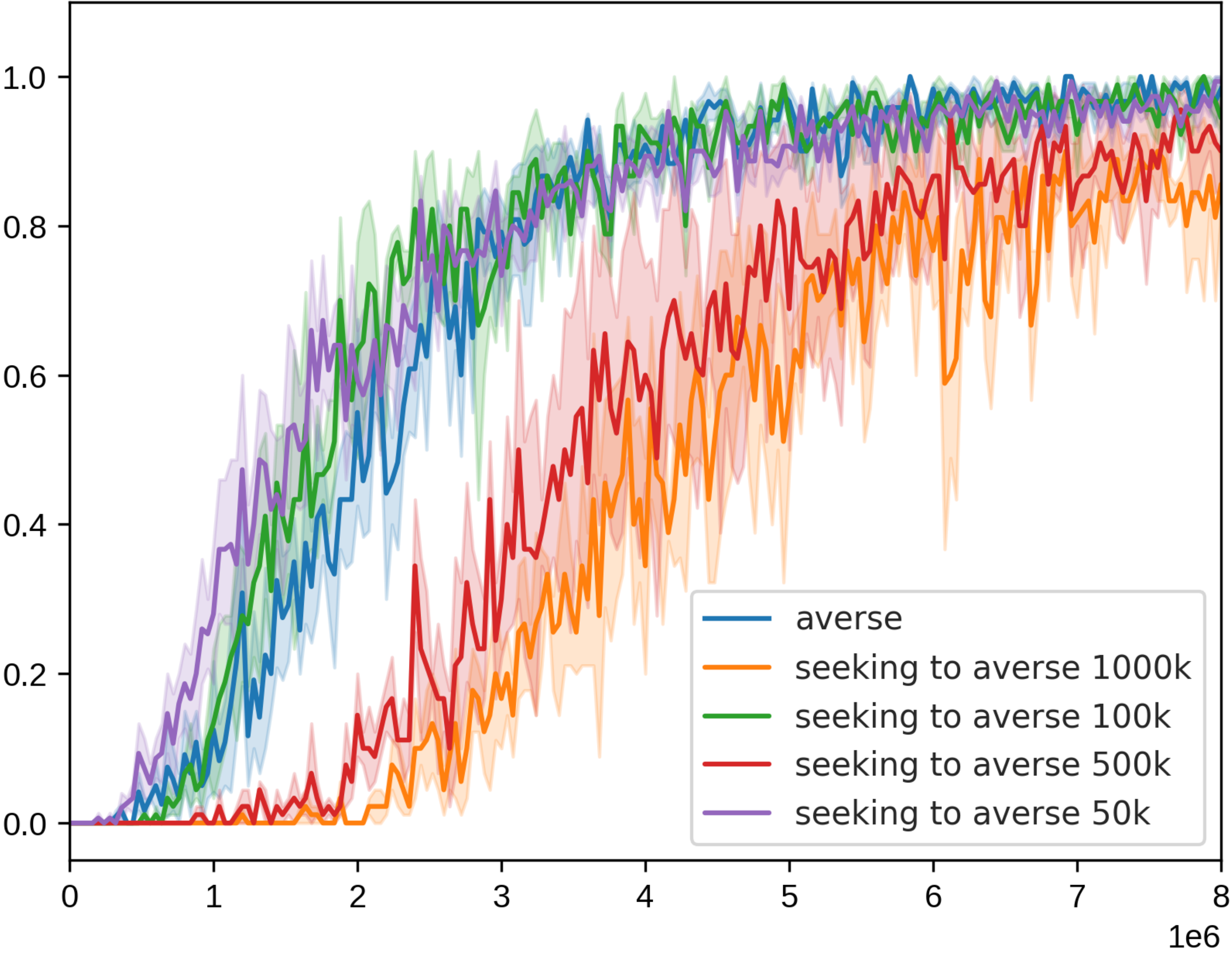}
  \end{center}
  \vspace{-0.60cm}
  \caption{Control scheduling steps in scenario \texttt{3s5z}}
  \vspace{-0.35cm}
  \label{sweep out}
\end{wrapfigure}

Training with static risk level makes the model experience only the particular risk trends even though the risk level is neutral. However, agents can efficiently explore behavior trends by scheduling risks to explore risk levels and optimistic behaviors at the beginning of learning. Especially, some \textit{super hard} scenarios show substantial performance improvements when scheduling risks as shown in \autoref{fig:results}, confirming that the effect of risk scheduling is significant where more exploration is required. In addition, we found that the longer the scheduling period is, the lower the performance, as shown in \autoref{sweep out}. Extending decaying steps result in the continued exploration rather than stable learning, similar to $\epsilon$-greedy scheduling. Therefore, appropriate scheduling steps are required.

%% file: Manuscript/5_Conclusion.tex
\section{Conclusion \& Future work}
\label{sec:conclusion}
In distributional reinforcement learning, we propose risk perspective exploration via risk scheduling. Scheduling risk level from a risk-seeking level to a specified risk level significantly accelerates learning speed and improves final performance in comparison to not scheduling it. In addition, the risk level is highly correlated with the challenging skills shown in SMAC scenarios, making the training feasible and stable. In the near future, we hope to exhibit this intriguing demonstration using various distributional algorithms that can handle single or multiple agents, as well as in various simulators.



%% file: Manuscript/6_Acknowledgements.tex
\section*{Acknowledgements}

This work was conducted by Center for Applied Research in Artificial Intelligence(CARAI) grant funded by Defense Acquisition Program Administration(DAPA) and Agency for Defense Development(ADD) (UD190031RD).

%% file: Appendix/1_Distributional_Reinforcement_Learning.tex
\section{Distributional Reinforcement Learning}
\begin{wrapfigure}{r}{0.40\columnwidth}
  \begin{center}
    \vspace{-1.0cm}
    \includegraphics[width=0.40\columnwidth]{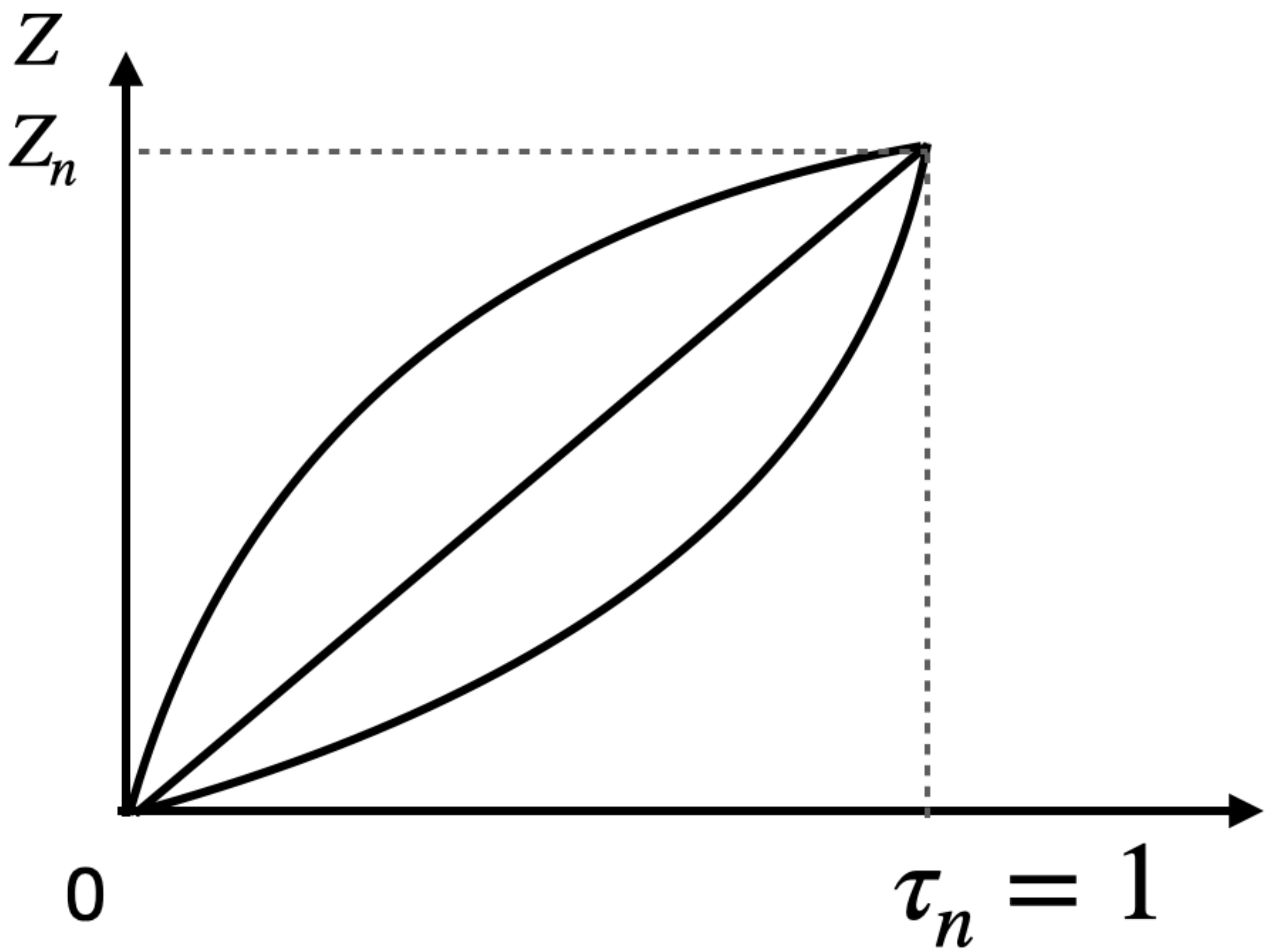}
  \end{center}
  \vspace{-0.40cm}
  \caption{Quantile regression}
  \vspace{-0.35cm}
  \label{quantile regression}
\end{wrapfigure}
Distributioanl deep RL got popular from \cite{bellemare2017distributional}. The output of distributional RL is a distribution of returns for a given state and action. They use the Wasserstein metric to calculate the TD-error between the distributional bellman updated distribution and the current distribution of returns in this approach. Although it differs across methods, most of them employ the quantile function to estimate the distribution of returns in order to calculate the Wasserstein distance between the current and objective distributions. Assume we have a model that approximates a quantile function with a domain of [0, 1] and a return range of ($-\infty, +\infty$). The output of a model may approximate the inverse of a cumulative density function like \autoref{quantile regression}.

\subsection{Algorithm}

\paragraph{DFAC} DFAC\citep{sun2021dfac} method, which is based on the IQN\cite{dabney2018implicit} algorithm, is the first to integrate distributional RL and multi-agent RL. IQN was utilized to sample quantile fractions from $\mathcal{U}[0, 1]$ for distributional output. The authors incorporated a distributional viewpoint in a multi-agent context using mean-shape decomposition without breaking the IGM condition, which can be stated as:

\begin{equation}
    \begin{aligned}
    \label{dfac_igm}
        \mathrm{arg\,max}_{\textbf{u}}\mathbb{E}[Z_{joint}(\textbf{h, u})] = \begin{pmatrix}
        \mathrm{arg\,max_{u_{1}}} \mathbb{E}[Z_{1}(\mathrm{h_{1}, u_{1}})]\\
        \vdots \\
        \mathrm{arg\,max_{u_{N}}} \mathbb{E}[Z_{N}(\mathrm{h_{N}, u_{N}})]\\
        \end{pmatrix}    
    \end{aligned}
\end{equation}
that can be proved by the following DFAC Theorem \citep{sun2021dfac}:
\begin{equation}
    \begin{aligned}
    \label{eqn:mean-shape}
        Z_{joint}\left (\bf{h}, \textbf{u} \right) &= \mathbb{E}[Z_{joint}\left (\textbf{h}, \textbf{u})\right] + Z_{joint}\left (\textbf{h}, \textbf{u} \right) - \mathbb{E}[Z_{joint}\left (\textbf{h}, \textbf{u})\right] \\
        &= Z_{mean}\left (\textbf{h}, \textbf{u}\right) + Z_{shape}\left (\textbf{h}, \textbf{u}\right) \\
        &= \psi ( Q_{1} \left (\mathrm{h}_{1}, \mathrm{u}_{1}\right), \dots, Q_{N}\left(\mathrm{h}_{N}, \mathrm{u}_{N} \right) )\\
        &+ \Phi ( Z_{1} \left (\mathrm{h}_{1}, \mathrm{u}_{1}\right), \dots, Z_{N}\left(\mathrm{h}_{N}, \mathrm{u}_{N} \right) )
    \end{aligned}
\end{equation}

which is proven to meet the IGM condition. DFAC outperforms all other algorithms, especially in difficult scenarios. This approach may also be modified to work with IQL, VDN\cite{sunehag2017value}, and QMIX\cite{rashid2018qmix}. The DMIX variations of the DFAC algorithm, which combines with QMIX, are employed as our baseline.

\paragraph{Exploration of DMIX} 
DMIX \cite{sun2021dfac} is another value-based RL version incorporating $\epsilon$-greedy exploration. However, this method leverages variance information from the return distribution while choosing an action. DMIX generates a return distribution using uniformly sampled quantile fractions from $\mathcal{U}[0, 1]$. If we set the number of quantile fractions to a tiny number, such as a digit of one, uniformly sampling quantile fractions stochastically results in exploiting tail information of a return distribution. For default, DMIX sets the number of quantile fractions to 8. As a result, this approach indirectly employs the variance of the return distribution per action for exploration.